\documentclass[10pt,twocolumn,letterpaper]{article}

\usepackage{cvpr}
\usepackage{times}
\usepackage{epsfig}
\usepackage{graphicx}
\usepackage{amsmath}
\usepackage{amssymb}
\usepackage{subfig}
\usepackage{multirow}
\usepackage{array}
\usepackage{booktabs}
\DeclareMathOperator{\IoU}{IoU}
\DeclareMathOperator{\Area}{area}
\DeclareMathOperator{\ArIoU}{ArIoU}
\DeclareMathOperator{\funL}{L}
\usepackage[breaklinks=true,bookmarks=false]{hyperref}

\cvprfinalcopy 

\def\cvprPaperID{3619} 

\begin{document}

\title{Learning a Rotation Invariant Detector with Rotatable Bounding Box}

\author{Lei Liu, Zongxu Pan, Bin Lei\\
Institute of Electronics, Chinese Academy of Sciences\\
{\tt\small \{lliu1, zxpan, leibin\}@mail.ie.ac.cn}
}

\maketitle

\begin{abstract}
   Detection of arbitrarily rotated objects is a challenging task due to the difficulties of locating the multi-angle objects and separating them effectively from the background. The existing methods are not robust to angle varies of the objects because of the use of traditional bounding box, which is a rotation variant structure for locating rotated objects. In this article, a new detection method is proposed which applies the newly defined rotatable bounding box (RBox). The proposed detector (DRBox) can effectively handle the situation where the orientation angles of the objects are arbitrary. The training of DRBox forces the detection networks to learn the correct orientation angle of the objects, so that the rotation invariant property can be achieved. DRBox is tested to detect vehicles, ships and airplanes on satellite images, compared with Faster R-CNN and SSD, which are chosen as the benchmark of the traditional bounding box based methods. The results shows that DRBox performs much better than traditional bounding box based methods do on the given tasks, and is more robust against rotation of input image and target objects. Besides, results show that DRBox correctly outputs the orientation angles of the objects, which is very useful for locating multi-angle objects efficiently. The code and models are available at \url{https://github.com/liulei01/DRBox}.
\end{abstract}
\section{Introduction}
Object detection is one of the most challenging tasks in computer vision and has attracted a lot of attentions all the time. Most existing detection methods use bounding box to locate objects in images. The traditional bounding box is a rotation variant data structure, which becomes a shortcoming when the detector has to deal with orientation variations of target objects. This article discusses how to design and train a rotation invariant detector by introducing the rotatable bounding box (RBox). Unlike traditional bounding box (BBox) which is the circumscribed rectangle of a rotated object, RBox is defined to involve the orientation information into its data structure. The proposed detector (Detector using RBox, DRBox) is very suitable for detection tasks that the orientation angles of the objects are arbitrarily changed.

Rotation invariant property becomes important for detection when the viewpoint of the camera moves to the top of the object thus the orientations of the objects become arbitrarily. A typical application is the object detection task in aerial and satellite images. Hundreds of remote sensing satellites are launched into space each year and generate huge amounts of images. Object detection on these images is of great importance, whereas the existing detection methods suffer from the lack of rotation invariant property. In the next paragraph, we firstly give a brief overview of the object detection methods, and then discuss how rotation invariant is taken into consideration by the recent methods.

Previous object detection methods usually use economic features and inference schemes for efficiency, and prevalent such methods include deformable part model (DPM)~\cite{felzenszwalb2010object}, selective search (SS)~\cite{uijlings2013selective} and EdgeBoxes~\cite{zitnick2014edge}. As the development of deep learning technique, deep neural networks (DNNs) have been applied for solving the object detection problem and the DNN based approaches achieve state-of-the-art detected performance. Among DNN based methods, region-based detector is widely used. Region-based convolutional neural network method (R-CNN)~\cite{girshick2014rich} makes use of SS for generating region proposals, and convolutional neural networks (CNN) is then employed on each proposal for detection. R-CNN is slow, in part because every proposal has to be fed into the network individually, in part because it is a multi-stage pipeline approach. Consequent approaches take much effort to integrate the detection pipelines gradually. Spatial pyramid pooling networks (SPPnets)~\cite{he2014spatial} speeds up R-CNN by sharing computation. The spatial pyramid pooling is introduced to remove the fixed size input constraint and the whole image needs only pass the net once. Several innovations are employed in Fast R-CNN~\cite{girshick2015fast} to improve both the effect and the efficiency of detection, including the use of RoI layer, multi-task loss, and the truncated SVD. Instead of using SS to generate region proposals, Faster R-CNN applies region proposal networks (RPNs) to generate the proposals~\cite{ren2015faster}, making all detection steps be integrated in an unified network, which achieves the best detected result at that time. Anchor boxes with multiple scales and aspect ratios are of the essence in Faster R-CNN, which locate the position of candidate objects. In Fast/Faster R-CNN, the network can be partitioned into a fully convolutional subnetwork which is independent of proposals and shares the computation, and a per-proposal subnetwork without sharing the computation. It is noticed that the per-proposal subnetwork is inefficient, and a region-based fully convolutional network (R-FCN) is proposed in~\cite{dai2016r} to remedy that issue by using FCN to share almost all computation on the entire image. By adding a branch to predict the mask of objects upon Faster R-CNN, Mask R-CNN can simultaneously detect the object and generate the segmentation mask of the object~\cite{he2017mask}. There is another kind of DL-based object detection approach that does not rely on region proposals, such as you only look once detector (YOLO)~\cite{redmon2016you} and single shot multibox detector (SSD)~\cite{liu2016ssd}. We refer to YOLO and SSD as box-based methods since they generate several boxes for detecting objects in the image according to certain rules. These boxes are called as prior boxes and objects are supposed to locate in or near certain prior box. SSD attempts to use pyramidal feature hierarchy computed from the convolutional net, however without reusing the higher resolution maps in the pyramidal feature hierarchy. The feature pyramid network (FPN)~\cite{lin2016feature} better utilizes the pyramidal feature hierarchy through introducing lateral connections which merge high-level semantic feature maps with coarse resolution and low-level semantic feature maps with refined resolution.

Many recent machine learning based methods have been used for remote sensing object detection. Many machine learning methods extract the candidate object feature, such as histogram of oriented gradients (HOG)~\cite{tuermer2013airborne,chen2016vehicle1}, texture~\cite{chen2016vehicle2}, bag-of-words (BoW)~\cite{sun2012automatic}, regional covariance descriptor (RCD)~\cite{chen2017building}, and interest point~\cite{wan2017affine}, followed by certain classifier, for example, sparse representation classifier (SRC)~\cite{chen2016vehicle1, chen2016vehicle2} and support vector machine (SVM)~\cite{chen2017building}. As the development of deep learning technique, DNNs have been successfully employed to solve object detection problem in remote sensing images. A CNN based pioneering study upon vehicle detection in satellite images is presented in\cite{chen2014vehicle}, followed by many approaches that focus on different types of geospatial object detection problems~\cite{jiang2015deep,vsevo2016convolutional,sommer2017fast,cheng2016learning,zhang2016weakly,long2017accurate}. As another effective feature extraction tool, deep Boltzmann machine is also used as feature extractor in geo-spatial object detection problems~\cite{han2015object,diao2016efficient}. 

To make the approach insensitive to objects’ in-plane rotation, some efforts are made either adjusting the orientation, or trying to extract rotation insensitive features. For example, to deal with the rotation variation of geo-spatial objects issue, a rotation invariant regularization term is introduced which enforces the samples and their rotated versions share the similar features~\cite{cheng2016learning}. Unlike these methods which try to eliminate the effect of rotation on the feature level, we prefer to make the rotation information useful for feature extraction so that the detection results involve the angle information of the objects. Therefore, the detection results is “rotatable”, whereas the performance of the detector is rotation invariant.

Zhou \etal proposed a network named ORN~\cite{zhou2017orn} which shows similar character with our method. ORN is used to classify images while extracting the orientation angle of the whole image. However, ORN can not be applied straightforward as a detection network, which needs to detect orientation locally on each object. In our method, the angle estimation is associated to plenty of prior boxes, so that the rotation of an object can be realized by the corresponding prior boxes, while other prior boxes are still available for other objects. Additionally, our method effectively separates the object proposals with its background pixels, so the angle estimation can concentrate on the object without the interfere of background.

The next section explains why rotatable bounding box (RBox) is better than BBox for rotation invariant detection. In section 3 we discuss how RBox takes place of BBox to form the newly designed detection method, DRBox. DRBox is tested on ship, airplane and vehicle detection in remote sensing images, exhibiting definitely superiority compared with traditional BBox based methods. 

\section{Rotatable bounding box}
Most object Detection methods use bounding box (BBox) to locate target objects in images. A BBox is a rectangle parameterized by four variables: the center point position (two variables), the width, and the height. BBox meets difficulties locating objects with different orientation angles. In this situation, BBox cannot provide the exact sizes of objects, and is very difficult to distinguish dense objects. Table~\ref{tab:rbox} lists the three main disadvantages of BBox and their corresponding examples. As shown in this table, a ship target in remote sensing image is aligned to BBoxes with different sizes and aspect ratios because of rotation.  As the result, the width and height of BBox have no relationship with the physical size of the rotated target. Another significant disadvantage is the discrimination between object and background. In the given example, about 60$\%$ region inside the BBox belongs to background pixels when the orientation angle of the ship is near 45 degrees. The situation becomes more difficult when target objects are distributed dense, in which case the objects are hard to be separated by BBoxes.

In this article, RBox is defined to overcome the above difficulties. RBox is a rectangle with a angle parameter to define its orientation. An RBox needs five parameters to define its location, size and orientation. Compared with BBox, RBox surrounds the outline of the target object more tightly, therefore overcomes all the disadvantages listed in the table. A detail comparison of RBox and BBox is demonstrated in Table~\ref{tab:rbox}. So we suggest that RBox is a better choice on detection of rotated objects.


\begin{table}
\begin{center}
\newcommand{\tabincell}[3]
{\begin{tabular}{@{}#1@{}}#2\end{tabular}}
\begin{tabular}{|p{0.28\linewidth}|p{0.28\linewidth}|p{0.28\linewidth}|}
\hline
\multicolumn{3}{|c|}{\textbf{Disadvantages of the traditional bounding box}}\\
\hline\hline
1. The size and aspect ratios do not reflect the real shape of the target object. & 2. Object and background pixels are not effectively separated. & 3. Dense objects are difficult to be separated.\\
\includegraphics[width=1\linewidth,height=0.9\linewidth]{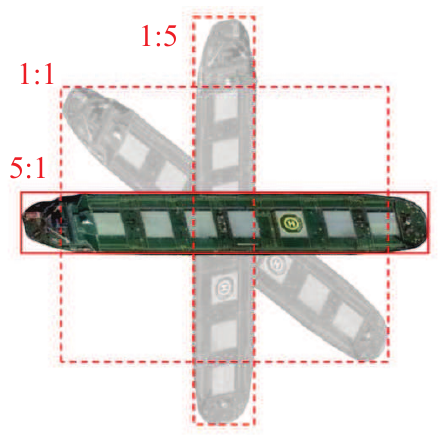}
&\includegraphics[width=1\linewidth,height=0.9\linewidth]{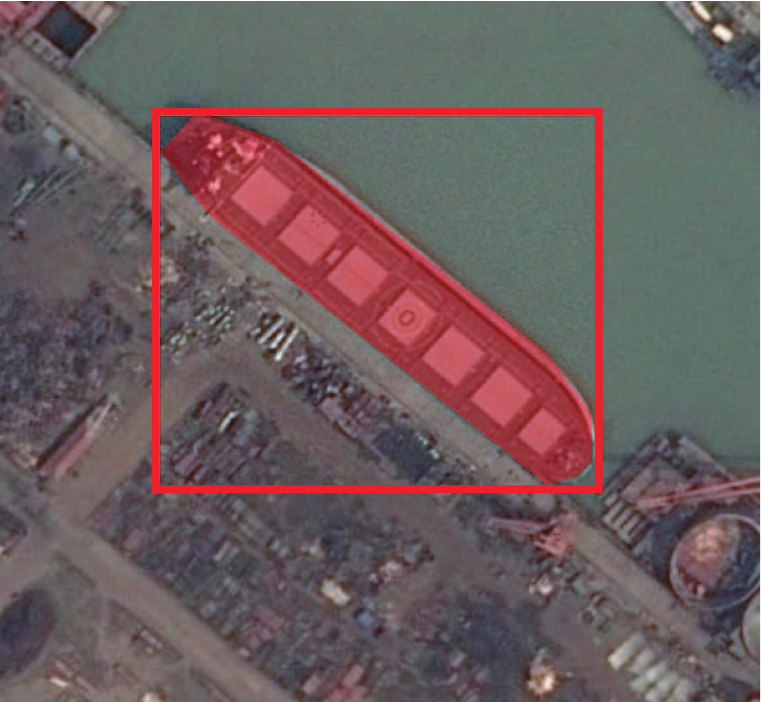}
&\includegraphics[width=1\linewidth,height=0.9\linewidth]{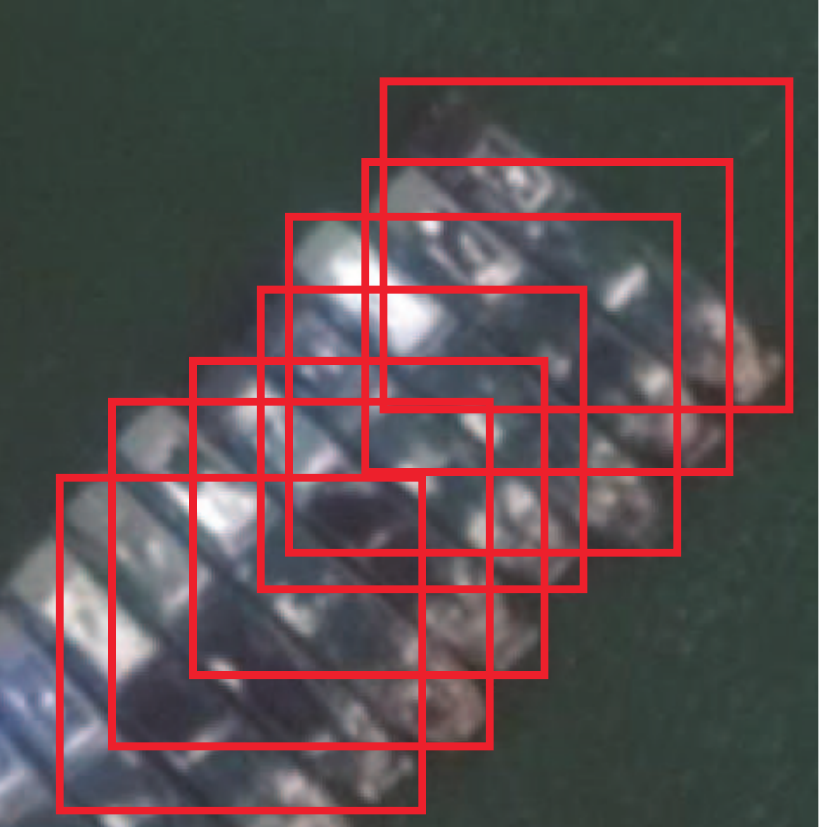}\\
\hline\hline
\multicolumn{3}{|c|}{\textbf{Advantages of the rotatable bounding box}}\\
\hline\hline
1. The width and height of RBox reflect the physical size of the object, which is helpful for customized designing of the prior boxes. & 2. RBox contains less background pixels than BBox does, so classification between object and background is easier. & 3. RBox can efficiently separate dense objects with no overlapped areas between nearby targets.\\
\includegraphics[width=1\linewidth,height=0.9\linewidth]{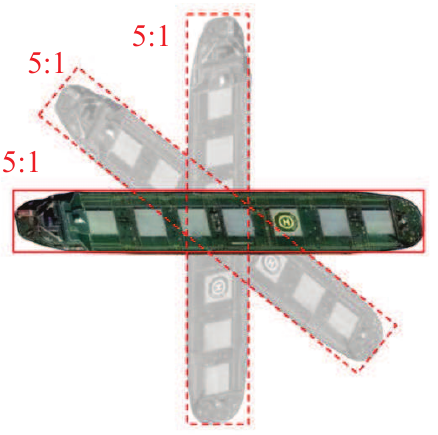}
&\includegraphics[width=1\linewidth,height=0.9\linewidth]{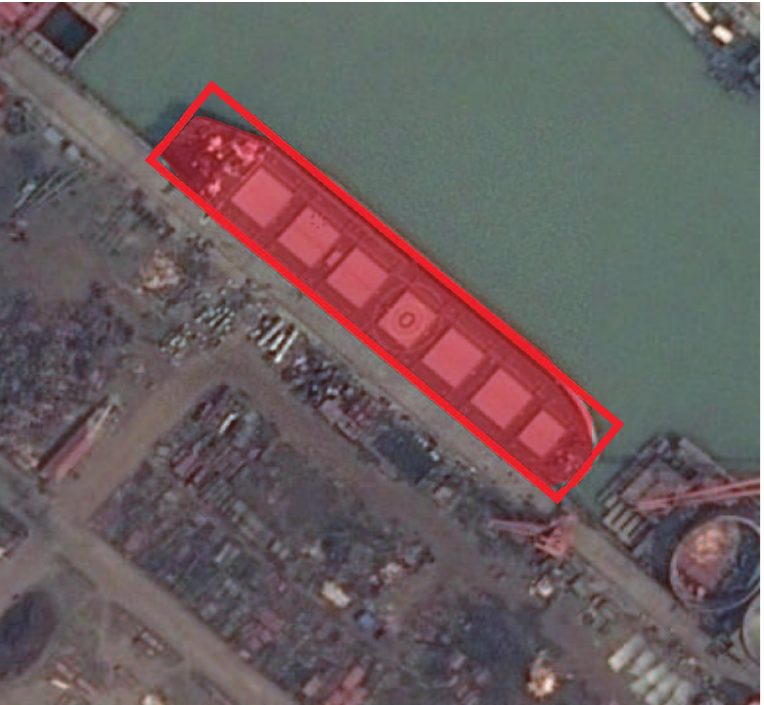}
&\includegraphics[width=1\linewidth,height=0.9\linewidth]{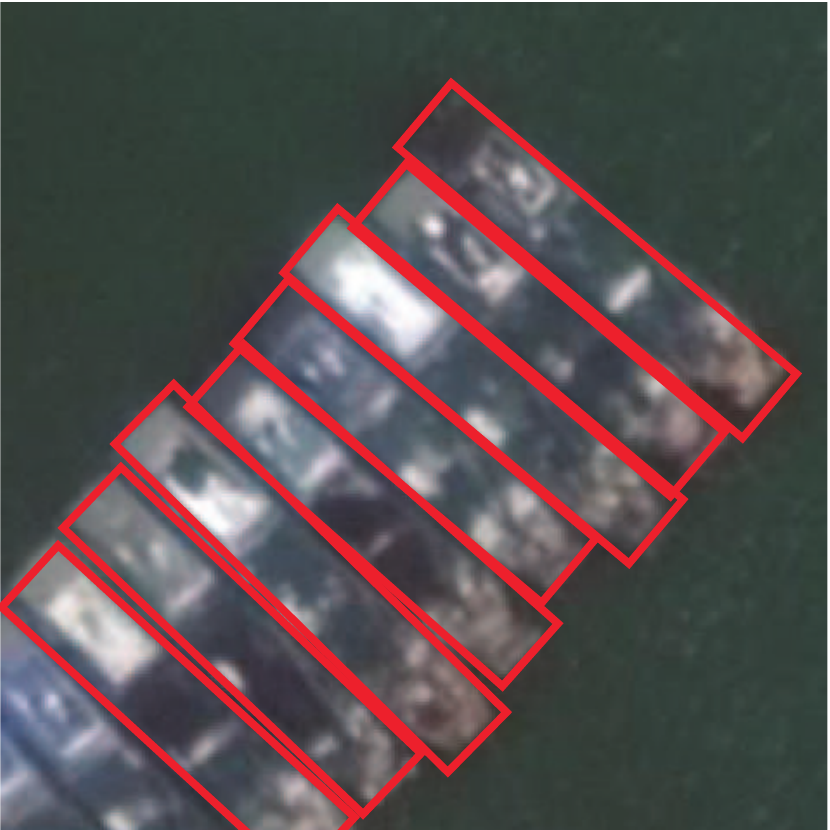}\\
\hline
\end{tabular}
\end{center}
\caption{Comparison of traditional bounding box and rotatable bounding box. Examples on ship locating are used to support the conclusions.}
\label{tab:rbox}
\end{table}

Given two boxes, it is important for a detection algorithm to evaluate their distance, which is used to select positive samples during training, and suppress repeated predictions in detection. The common used criterion for BBox is the Intersection-over-Union (IoU), which can also be used by RBox. The IoU between two RBoxes $A$ and $B$ defines as following:
\begin{equation} \label {eq:iou}
\IoU\left( {A,B} \right) = \frac{{\Area\left( {A \cap B} \right)}}{{\Area\left( {A \cup B} \right)}},
\end{equation}
where $\cap$ and $\cup$  are Boolean operations between two RBoxes. The Boolean calculation between RBoxes is more complex than BBox because the intersection of two RBoxes can be any polygon with no more than eight sides.

Another criterion for RBox is angle-related IoU (ArIoU), which is defined as following: 
\begin{equation} \label {eq:ariou}
\ArIoU\left( {A,B} \right) = \frac{{\Area\left( {\hat A \cap B} \right)}}{{\Area\left( {\hat A \cup B} \right)}}\cos \left( {{\theta _A} - {\theta _B}} \right),
\end{equation}
or
\begin{equation} \label {eq:ariou180}
\ArIoU_{180}\left( {A,B} \right) = \frac{{\Area\left( {\hat A \cap B} \right)}}{{\Area\left( {\hat A \cup B} \right)}}\left| {\cos \left( {{\theta _A} - {\theta _B}} \right)} \right|,
\end{equation}
where $\theta_A$ and $\theta_B$ are angles of RBox $A$ and $B$, $\hat A$ is an RBox which keeps the same parameters with RBox $A$ except that the angle parameter is $\theta_B$, not $\theta_A$. ArIoU takes angle difference into consideration so that the ArIoU between RBox $A$ and $B$ decreases monotonically when their angle difference changes from 0 degree to 90 degrees. The two definitions differ in the behavior when $(\theta_A-\theta_B)$ is near 180 degrees. $\ArIoU_{180}$  ignores the head and tail direction of the objects when distinguish them is too difficult.

IoU and ArIoU are used in different way. ArIoU is used for training so it can enforce the detector to learn the right angle, while IoU is used for non-maximum suppression (NMS) so the predictions with inaccurate angle can be effectively removed.

\section{Rotation invariant detection}
In this section, we apply RBox into detection, which means that the detector must learn not only the locations and sizes, but also the angles of the target objects. Once this purpose is achieved, the network can “realize” the existence of orientation difference between objects, rather than being confused by rotation. The performance of the detector becomes rotation invariant as the result.

\subsection{Model}
\begin{figure}
\begin{center}
   \includegraphics[width=1.0\linewidth]{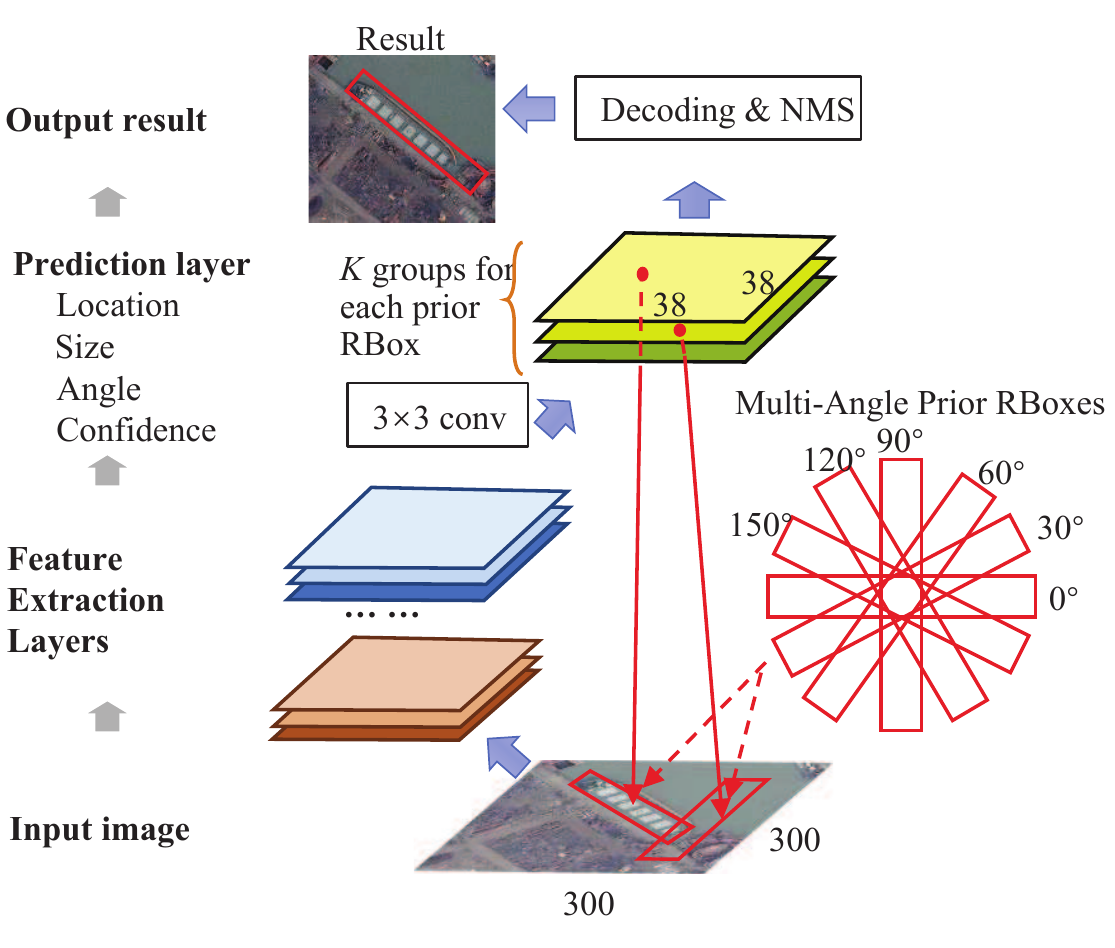}
\end{center}
   \caption{The networks structure of DRBox. The networks structure of DRBox is similar with other box based methods except for the use of multi-angle prior RBoxes. DRBox searches for objects using sliding and rotating prior RBoxes on input image and then output locations of objects besides with their orientation angles.}
\label{fig:network}
\end{figure}

NETWORK STRUCTURE: DRBox uses a convolutional structure for detection, as shown in Figure~\ref{fig:network}. The input image goes through multi-layer convolution networks to generate detection results. The last convolution layer is for prediction and the other convolution layers are for feature extraction. The prediction layer includes $K$ groups of channels where $K$ is the number of prior RBoxes in each position. Prior RBoxes is a series of predefined RBoxes. For each prior RBox, the prediction layer output a confidence prediction vector indicating whether it is a target object or background, and a 5 dimensional vector which is the offset of the parameters between the predicted RBox and the corresponding predefined prior RBox. A decoding process is necessary to transform the offsets to the exact predicted RBoxes. At last, the predicted RBoxes are sorted with their confidence and passed through NMS to remove repeated predictions.

\textbf{Multi-angle prior RBox} plays an important role in DRBox. The convolutional structure ensures that prior boxes can move over different locations to search for target objects. On each location, the prior RBoxes rotate at a series of angles to generate multi-angle predictions, which is the key difference between DRBox and other bounding box based methods. The aspect ratio used in detection is fixed according to the object type, which decreased the total number of prior boxes. By the multi-angle prior RBoxes strategy, the network is trained to treat the detection task as a series of sub-tasks. Each sub-task focuses on one narrow angle range, therefore decreases the difficulty caused by rotation of objects.

\subsection{Training}
The training of DRBox is extended from SSD training procedure~\cite{liu2016ssd} to involve angle estimation. During training, each ground truth RBox is assigned with several prior RBoxes according to their ArIoU. ArIoU is a non-commutative function, which means $\ArIoU({A,B})$ is different from $\ArIoU({B,A})$. A prior RBox $P$ is assigned to a ground truth RBox $G$ when $\ArIoU({P,G})>0.5$. After the assignment, the matched prior RBoxes are considered as positive samples and are responsible for generating the losses of location and angle regression. The use of ArIoU helps the training process to select prior RBox with proper angle as positive samples, so the angle information of the objects can be roughly learned during training. After the matching step, most of the prior RBoxes are negative. We apply hard negative mining to decrease the number of the negative samples.

The objective loss function of DRBox is extended from SSD objective loss function by adding the angle related term. The overall objective loss function is as following:
\begin{equation} \label {eq:l}
\funL\left( {x,c,l,g} \right) = \frac{1}{N}\left( {{\funL_{\text{conf}}}\left( c \right) + {\funL_{\text{rbox}}}\left( {x,l,g} \right)} \right)
\end{equation}
where $N$ is the number of matched prior RBoxes. The confidence loss $\funL_{\text{conf}}(c)$ is a two class softmax loss over all selected positive and negative samples, where $c$ is the two-dimension confidence vector. The RBox regression loss $\funL_{\text{rbox}}(x,l,g)$ is similar to SSD and Faster R-CNN, where we calculate the smooth $L_1$ loss between the predicted RBox $l$ and the ground truth RBox $g$:
\begin{equation} \label {eq:lrbox}
\begin{split}
&{\funL_{\text{rbox}}}\left( {x,l,g} \right)\\ 
&= \sum\limits_{i \in Pos} {\sum\limits_j {\sum\limits_{m \in \left\{ {cx,cy,w,h,a} \right\}} {{x_{ij}}\text{smooth}_{\text{L1}}}\left( {\hat l_i^m - \hat g_j^m} \right)} }  
\end{split}
\end{equation}
where $x_{ij}\in\{1,0\}$ is an indicator for matching the $i$-th prior RBox to the $j$-th ground truth RBox. $\hat l$ and $\hat g$ are defined as following, which are the offsets of the parameters in $l$ and $g$ with their corresponding prior RBox $p$, respectively:\begin{subequations}
\begin{equation} \label {eq:lhat1}
{\hat t^{cx}} = {\left( {{t^{cx}} - {p^{cx}}} \right)} \mathord{\left/
 {\vphantom {{\left( {{t^{cx}} - {p^{cx}}} \right)} {{p^w}}}} \right.
 \kern-\nulldelimiterspace} {{p^w}},  \quad
{\hat t^{cy}} = {{\left( {{t^{cy}} - {p^{cy}}} \right)} \mathord{\left/
 {\vphantom {{\left( {{t^{cy}} - {p^{cy}}} \right)} {{p^h}}}} \right.
 \kern-\nulldelimiterspace} {{p^h}}};
\end{equation}
\begin{equation}\label{eq:lhat2}
{\hat t^w} = \log \left( {{{{t^w}} \mathord{\left/
 {\vphantom {{{t^w}} {{p^w}}}} \right.
 \kern-\nulldelimiterspace} {{p^w}}}} \right), \quad
{\hat t^h} = \log \left( {{{{t^h}} \mathord{\left/
 {\vphantom {{{t^h}} {{p^h}}}} \right.
 \kern-\nulldelimiterspace} {{p^h}}}} \right);
\end{equation}
\begin{equation}\label{eq:lhat3}
{\hat t^a} = \tan \left( {{t^a} - {p^a}} \right).
\end{equation}
\end{subequations}

Equations~\ref{eq:lhat1} , ~\ref{eq:lhat2} and ~\ref{eq:lhat3} are the location regression terms, the size regression terms and \textbf{the angle regression term}, respectively. The angle regression term applies tangent function to adapt to the periodicity of the angle parameter. The minimization of the angle regression term ensures that the correct angle is learned during training.

\subsection{Complement details}
PYRAMID INPUT: DRBox applies pyramid input strategy that the original image is rescaled into different resolutions, and separated into overlapped $300\times300$ sub-images. The DRBox network is applied to each sub-image and the network only detects targets with proper size. Non-maximum suppression is applied on the detection results of the whole image, which suppresses repeated predictions not only within a sub-image, but also crossing overlap areas of different sub-images. The pyramid input strategy helps the detection network to share features between large and small objects. Besides, the satellite image used in this article is often very large, so the division process and non-maximum suppression across sub-images helps to detect objects in very large images.

CONVOLUTION ARCHITECTURE: DRBox uses truncated VGG-net for detection. All the full-connective layers, convolution layers and pooling layers after layer conv4\_3 are removed. Then, a $3\times3$ convolution layer is added after layer conv4\_3. The receptive field of DRBox is 108 pixels × 108 pixels, so any targets larger than this scope cannot be detected. Besides, the feature map of layer conv4\_3 is $38\times38$, so the targets closer than 8 pixels may be missed.

PRIOR RBOX SETTINGS: In this article, three DRBox  networks are trained separately for vehicle detection, ship detection and airplane detection, respectively. The scale and input resolution settings jointly ensure that the areas of the prior RBoxes cover the sizes of the objects sufficiently, thus the objects of different sizes can be effectively captured. In ship detection, it is hard to distinguish the head and tail of the target. In this case, the angles of the ground truth RBoxes and multi-angle prior RBoxes varies from 0 degree to 180 degrees. In detail, ship objects are detected with prior RBoxes of $20\times8$, $40\times14$, $60\times17$, $80\times20$, $100\times25$ pixels in size and 0:30:150 degrees in angle; vehicle objects are detected with prior RBoxes of $25\times9$ pixels in size and 0:30:330 degrees in angle; airplane objects are detected with prior RBoxes of $50\times50$, $70\times70$ pixels in size and 0:30:330 degrees in angle. The total number of prior boxes per image is 43320, 17328 and 34656 for ship, vehicle and airplane detection, respectively. 

DRBox reaches 70-80 fps on NVIDIA GTX 1080Ti and Intel Core i7. The input pyramid strategy produce no more than $4/3$ times time cost. Considering $1/3$ overlapped between sub-images, DRBox reaches processing a speed of $1600\times1600$ pixels$^{2}$ per second. The speed of SSD and Faster R-CNN are 70 fps and 20 fps on our dataset using the same convolution network architecture.  
\section{Experiments and results}
\subsection{Dataset}
We apply our method on object detection of satellite images. We have not found any open source dataset on this problem, so we build one using the GoogleEarth images. The dataset includes three categories of objects: vehicles, ships and airplanes. The vehicles are collected from urban area in Beijing, China. The ships are collected near the wharfs and ports besides the Changjiang River, the Zhujiang River and the East China Sea. The airplanes are collected from the images of 15 airports in China and America. The dataset recently includes about 12000 vehicles, 3000 ships and 2000 airplanes and is still under expansion. About 2000 vehicles, 1000 ships and 500 airplanes are taken out as testing dataset and others are used for training.

Each object in the images are marked with a RBox, which indicates not only the location and size, but also the angle of the object. A Matlab tool is developed to label the data with RBoxes.

\subsection{Benchmark}
In this section, we compare the performance of DRBox with the detectors that use BBox. SSD and Faster R-CNN are used as benchmarks of BBox based methods. All the detectors use the same convolution architecture and the same training data argumentation strategy. The prior boxes used in SSD and the anchor boxes used in Faster R-CNN are optimized for the datasets. All other hyper parameters are optimized for the dataset, too.

\subsection{Detection results}
\begin{figure*}
\begin{center}
   \subfloat[]{
     \includegraphics[width=0.18\linewidth,height=0.18\linewidth]{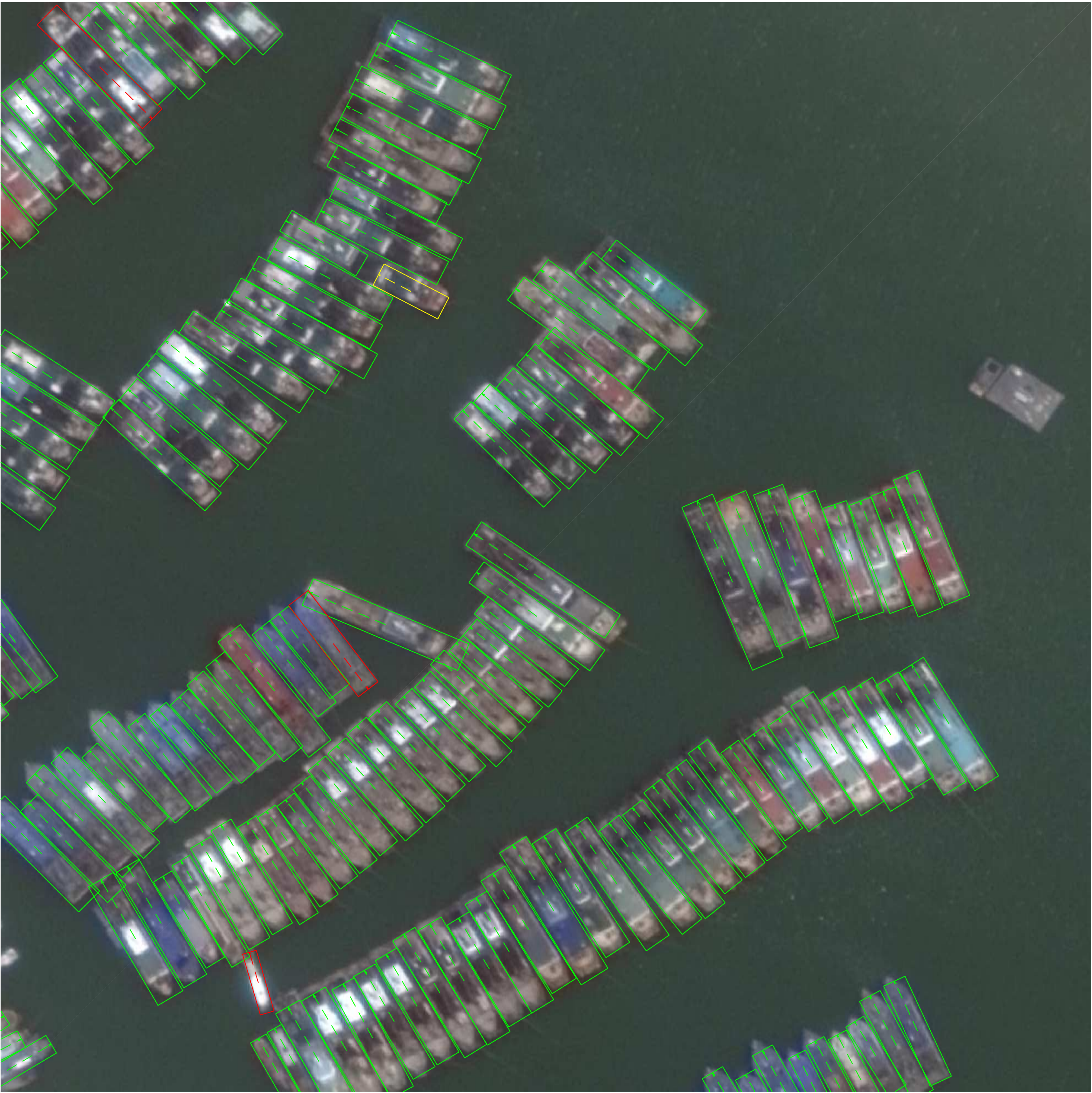}}
     \label{fig:comp1}
   \subfloat[]{
     \includegraphics[width=0.18\linewidth,height=0.18\linewidth]{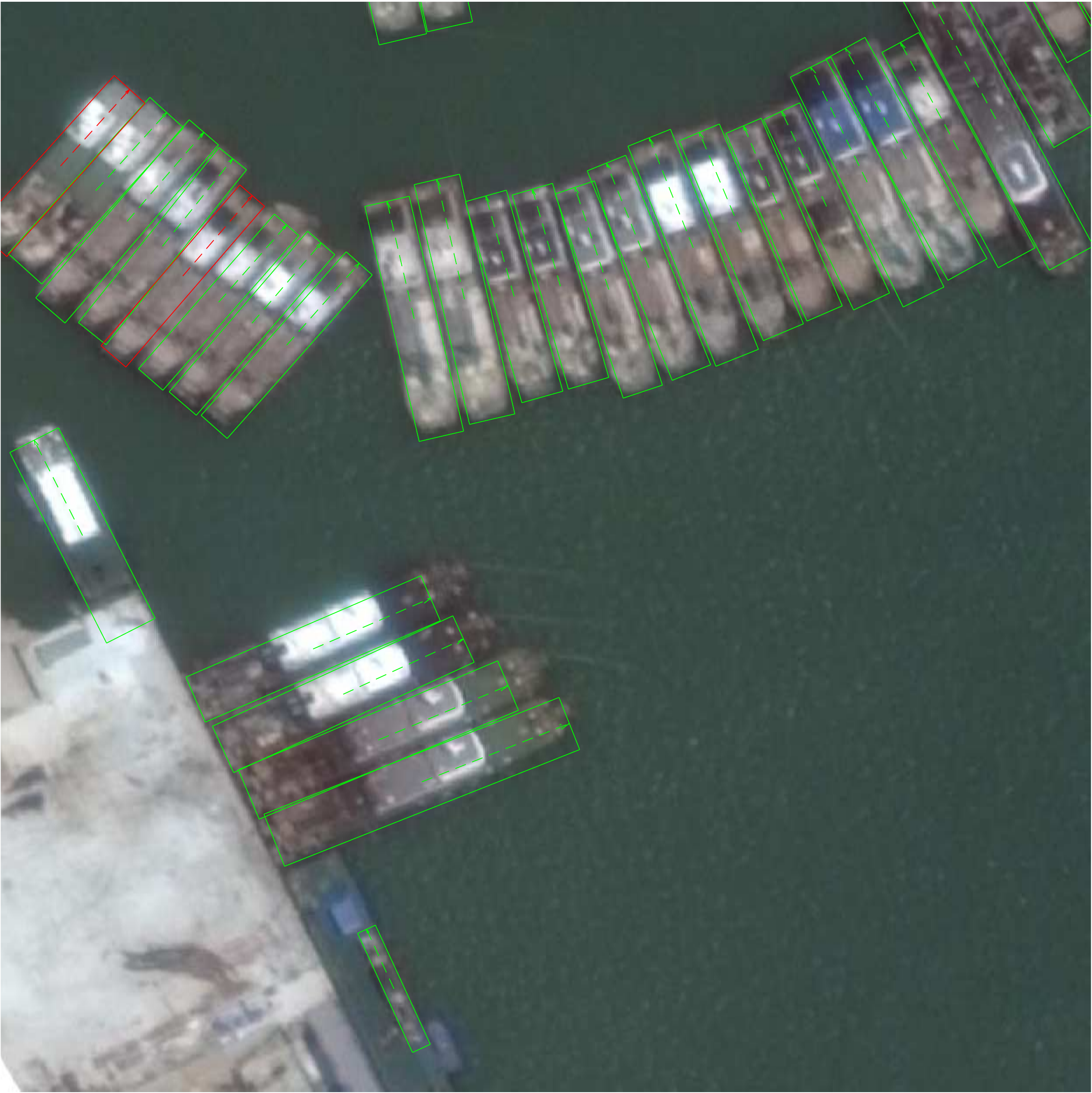}}
     \label{fig:comp2}
   \subfloat[]{
     \includegraphics[width=0.18\linewidth,height=0.18\linewidth]{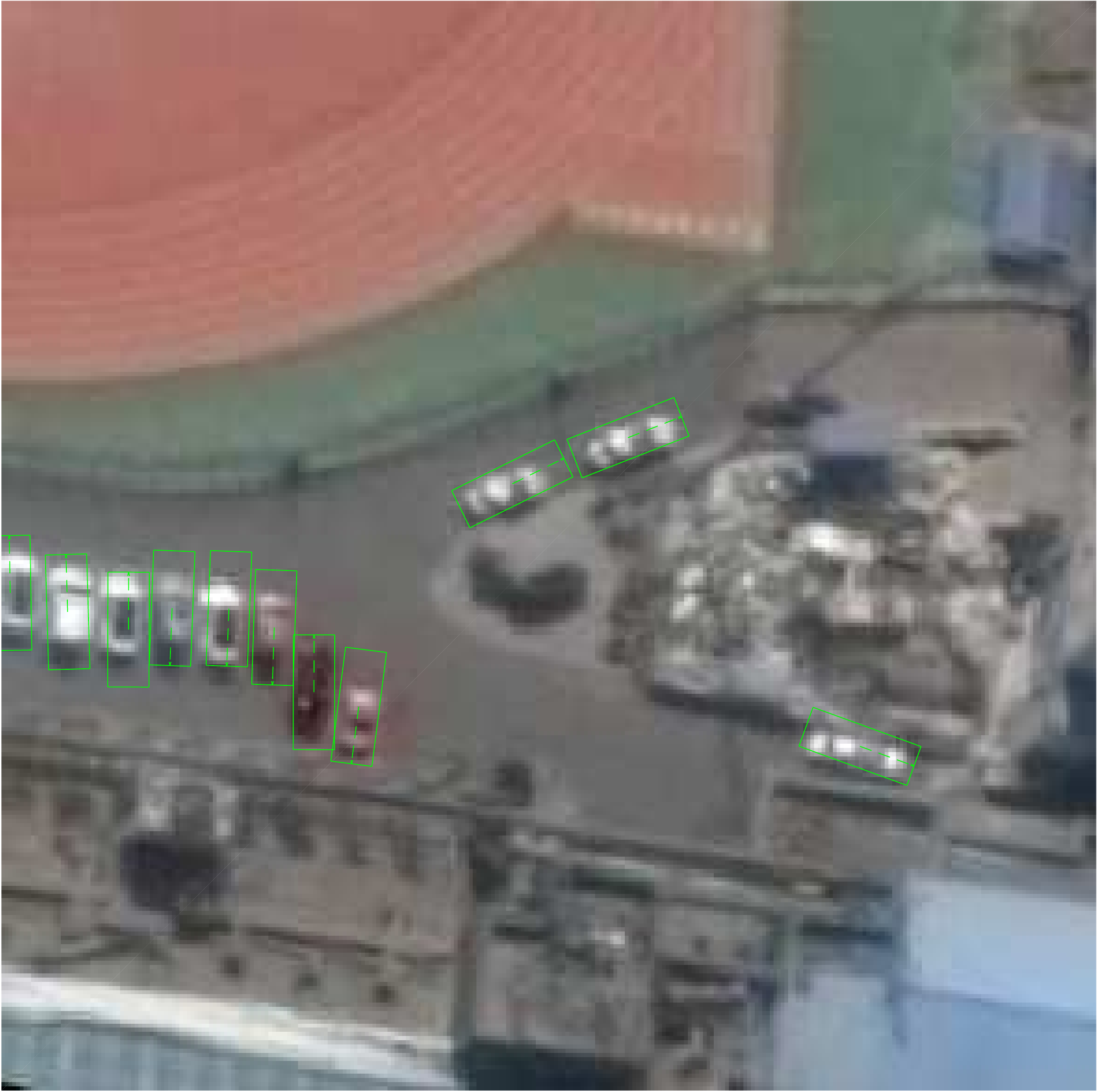}}
     \label{fig:comp3}
\subfloat[]{
     \includegraphics[width=0.18\linewidth,height=0.18\linewidth]{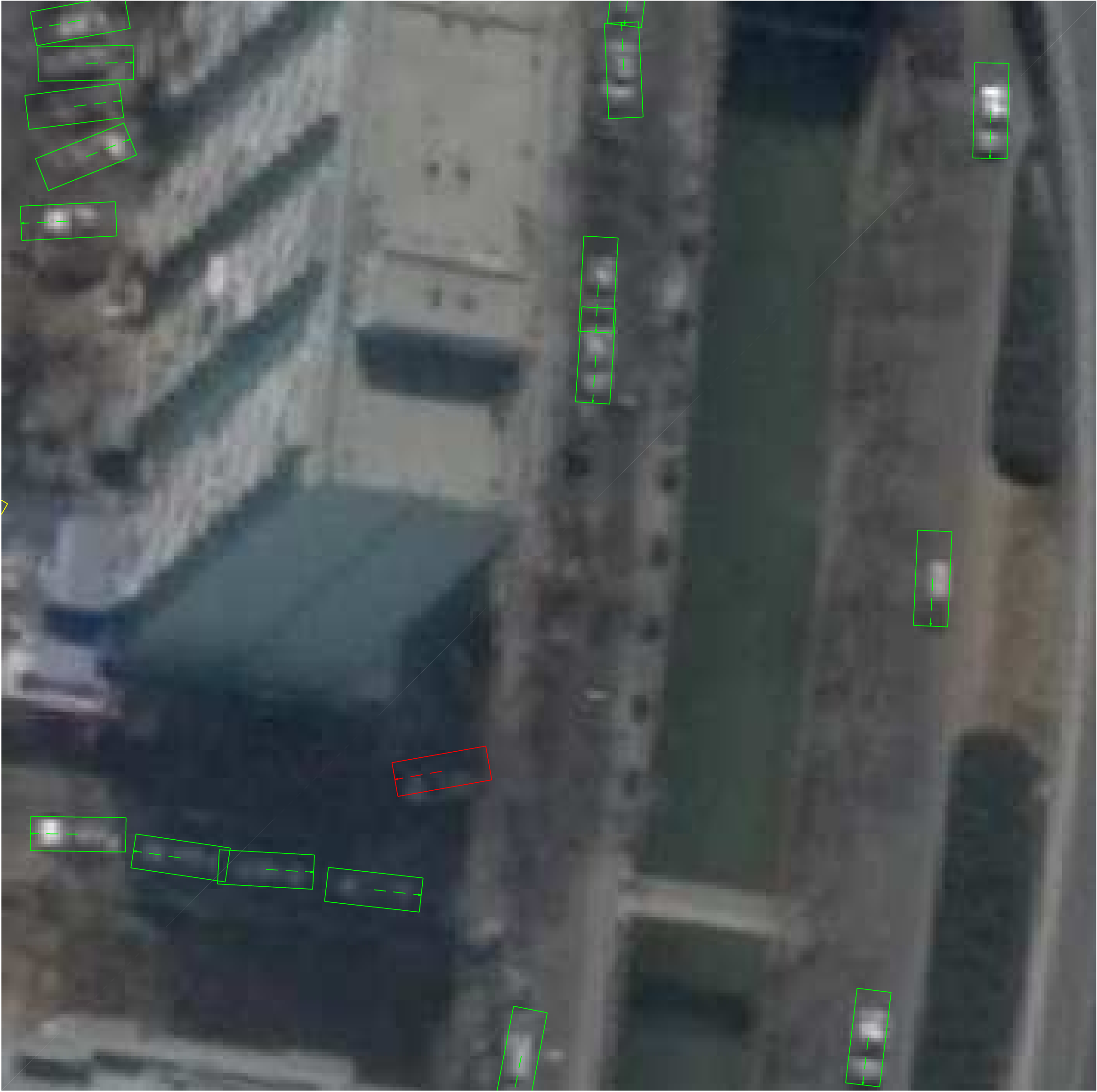}}
     \label{fig:comp4}
\subfloat[]{
     \includegraphics[width=0.18\linewidth,height=0.18\linewidth]{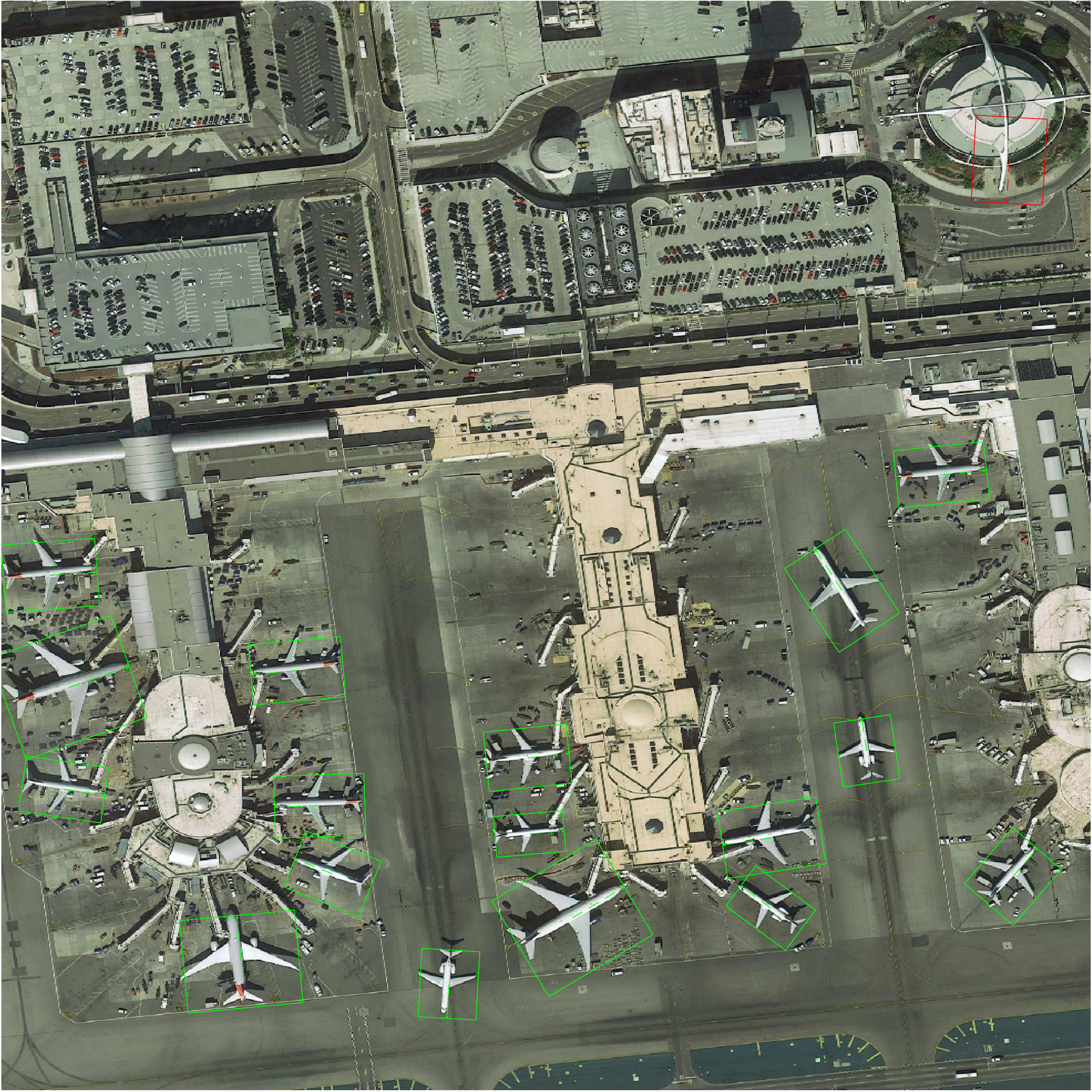}}
     \label{fig:comp5}
\subfloat[]{
     \includegraphics[width=0.18\linewidth,height=0.18\linewidth]{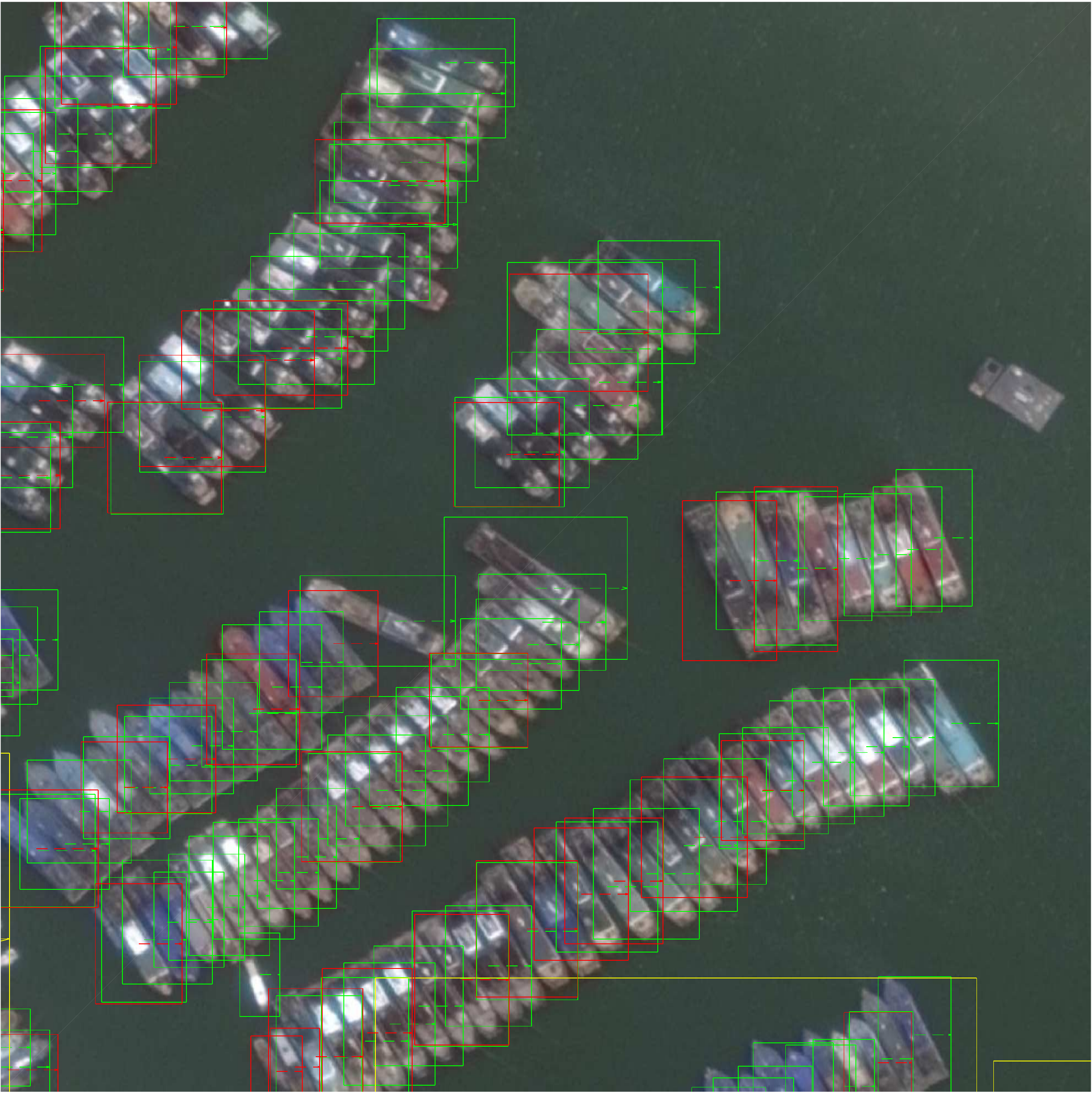}}
     \label{fig:comp6}
   \subfloat[]{
     \includegraphics[width=0.18\linewidth,height=0.18\linewidth]{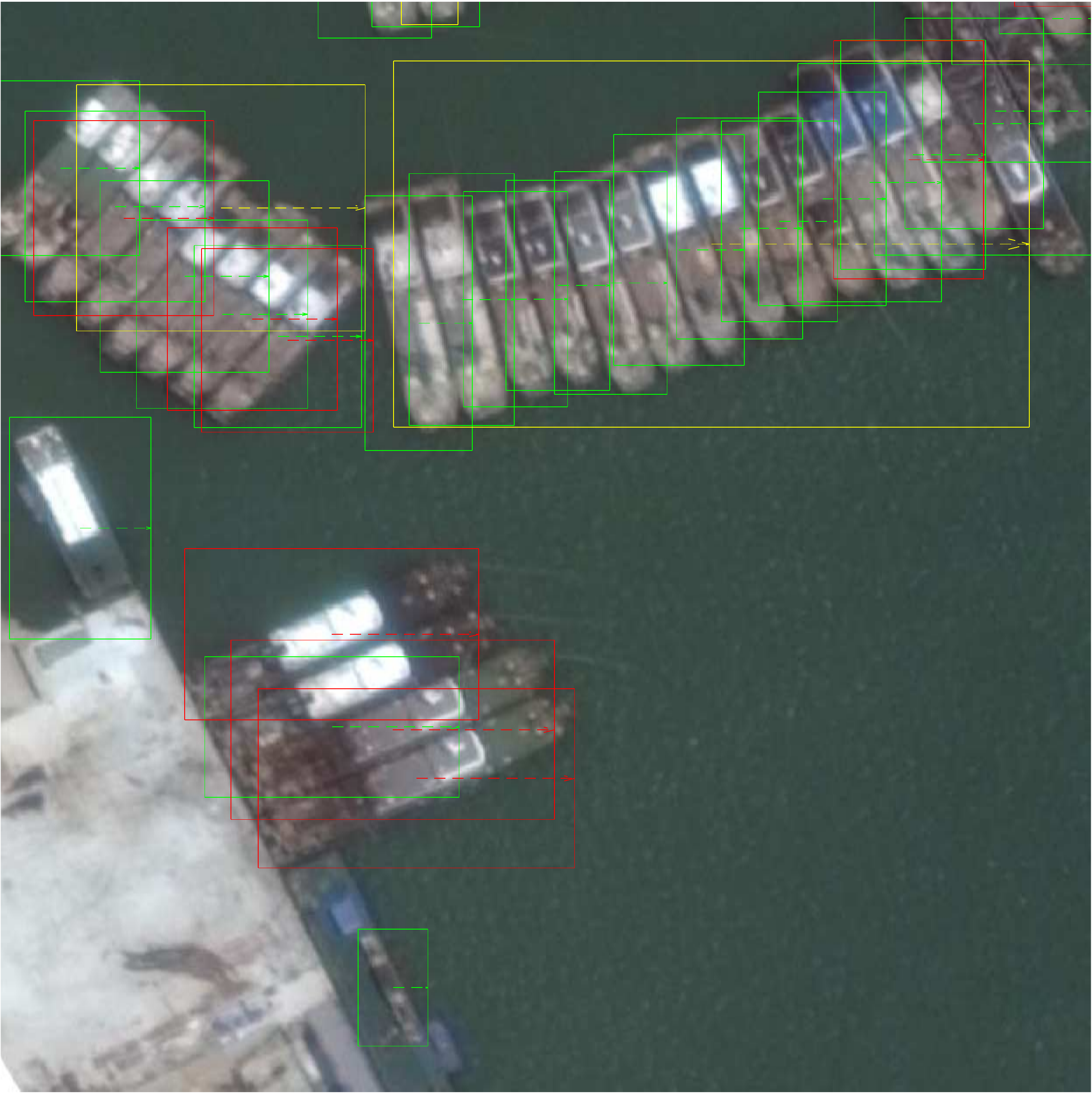}}
     \label{fig:comp7}
   \subfloat[]{
     \includegraphics[width=0.18\linewidth,height=0.18\linewidth]{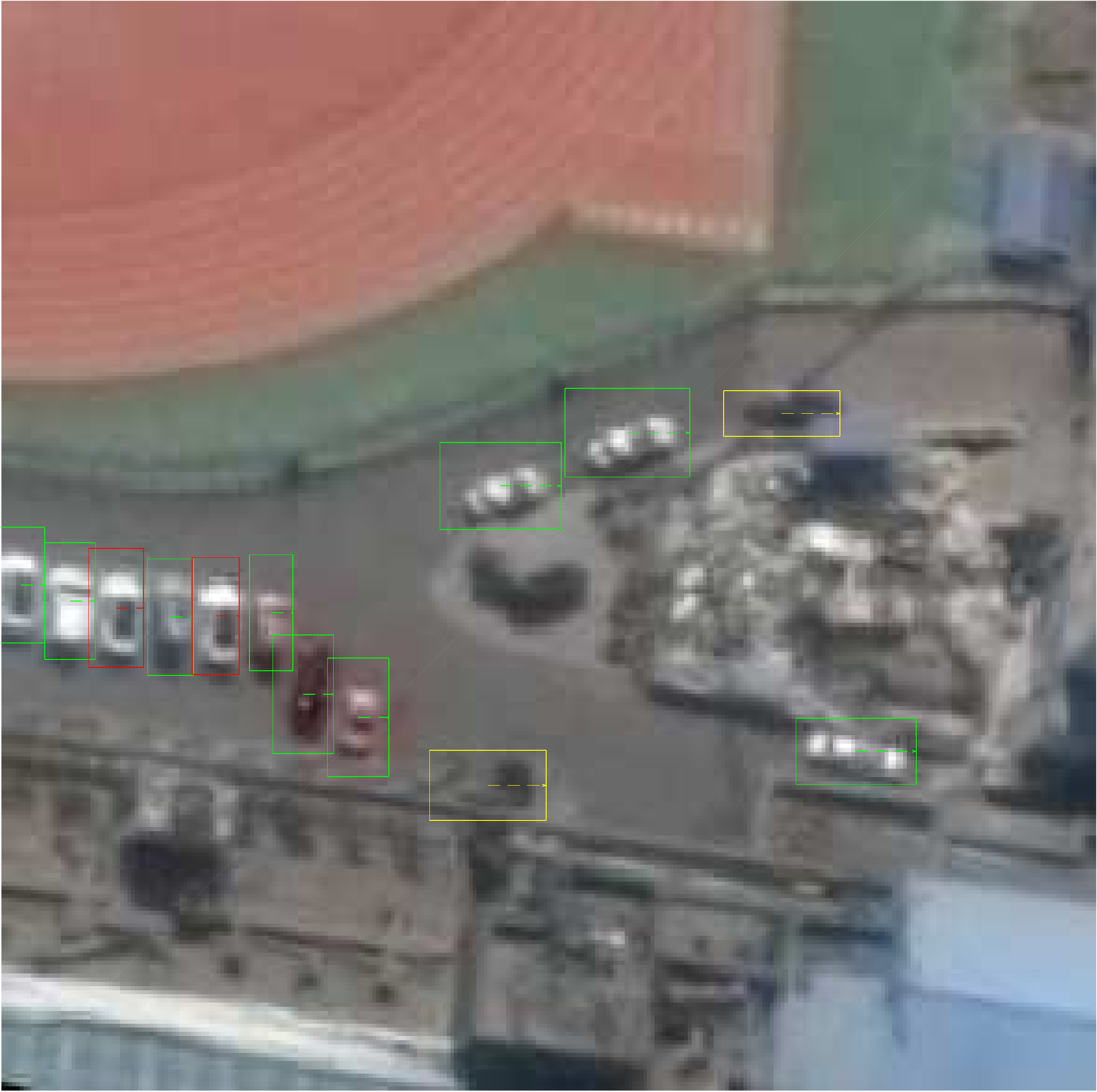}}
     \label{fig:comp8}
\subfloat[]{
     \includegraphics[width=0.18\linewidth,height=0.18\linewidth]{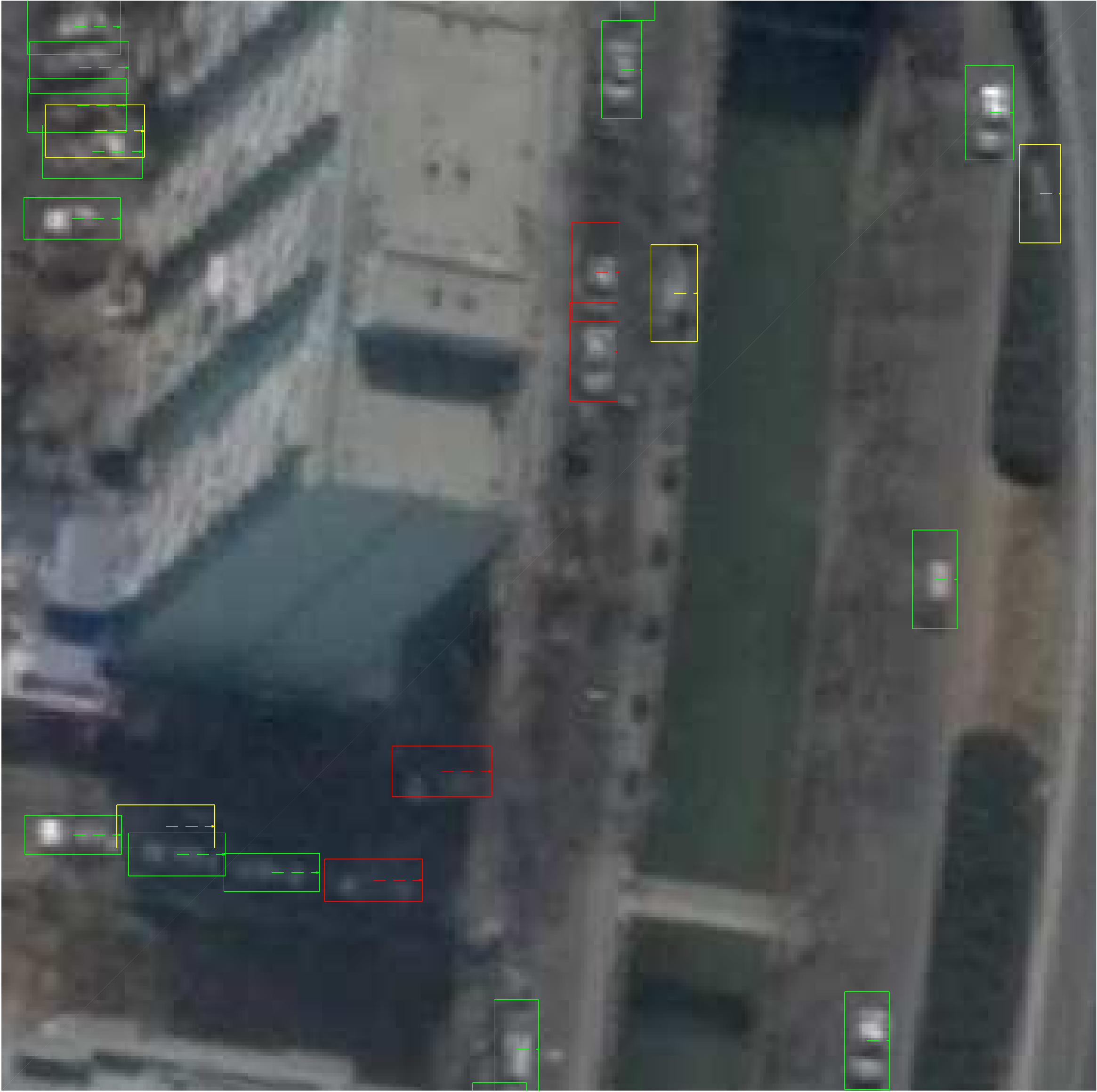}}
     \label{fig:comp9}
\subfloat[]{
     \includegraphics[width=0.18\linewidth,height=0.18\linewidth]{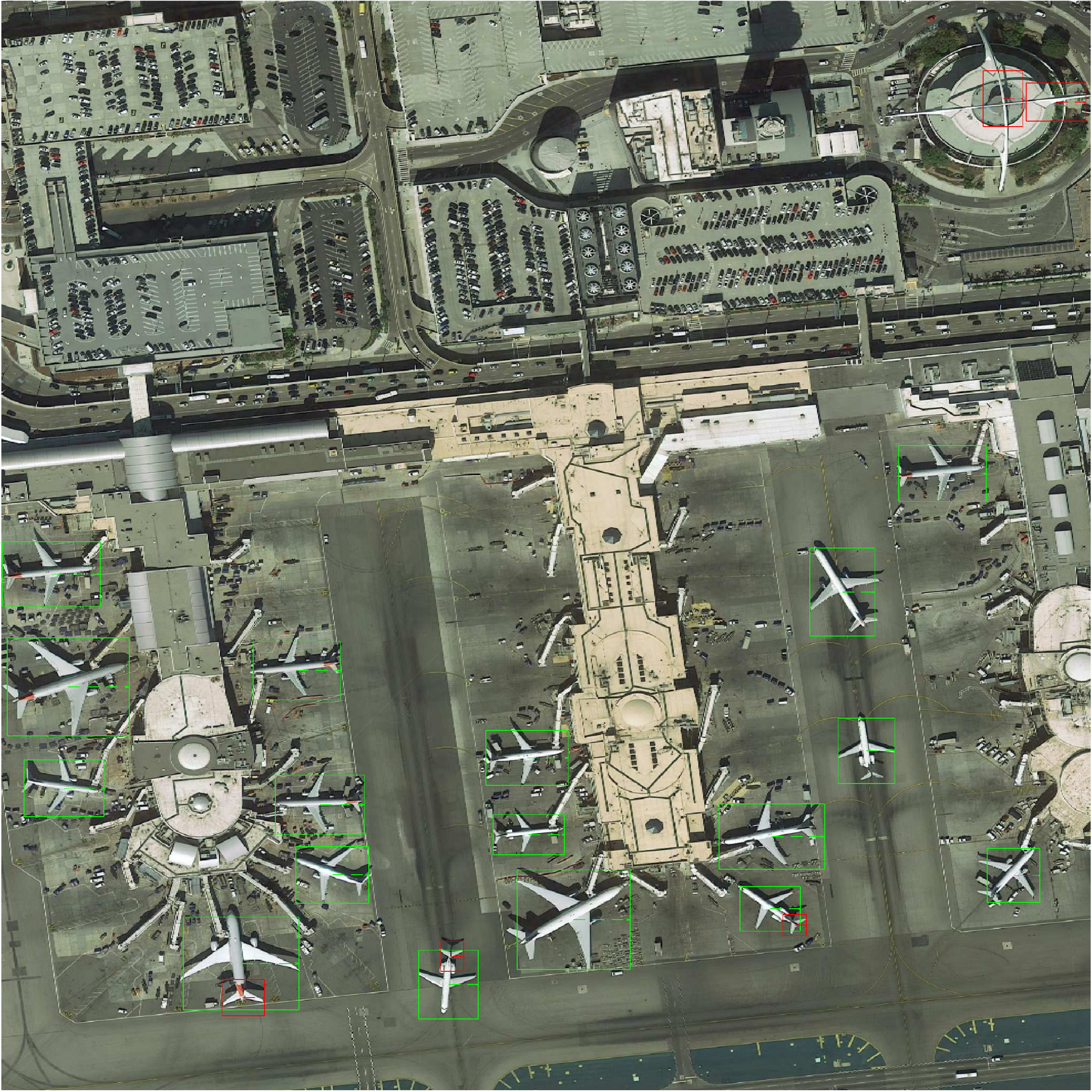}}
     \label{fig:comp10}
\end{center}
   \caption{Detection results using RBox and traditional bounding box. The first row are results of DRBox; the second row are results of SSD. Columns 1-2, columns 3-4 and column 5 are results of ship detection, vehicle detection and airplane detection, respectively. DRBox performs better in the given examples.}
\label{fig:comp}
\end{figure*}

Figure~\ref{fig:comp} show ships, vehicles and airplanes detected by using RBox (DRBox) and BBox (SSD). The predicted bounding boxes that matches the ground truth bounding boxes with $\text{IoU} > 0.5$ are plotted in green color, while the false positive predictions and false negative predictions are plotted in yellow and red color, respectively. Our method is better in the given scenes. DRBox successfully detects most of the ships on both the port region and open water region, while SSD almost fails on detection of nearby ships. The detection results for vehicles and airplanes also show that SSD generates more false alarms and false dismissals than does DRBox.  

\begin{figure*}
\begin{center}
   \subfloat[]{
     \includegraphics[width=0.18\linewidth,height=0.18\linewidth,trim={6.768cm 2.127cm 6.111cm 11.6cm },clip]{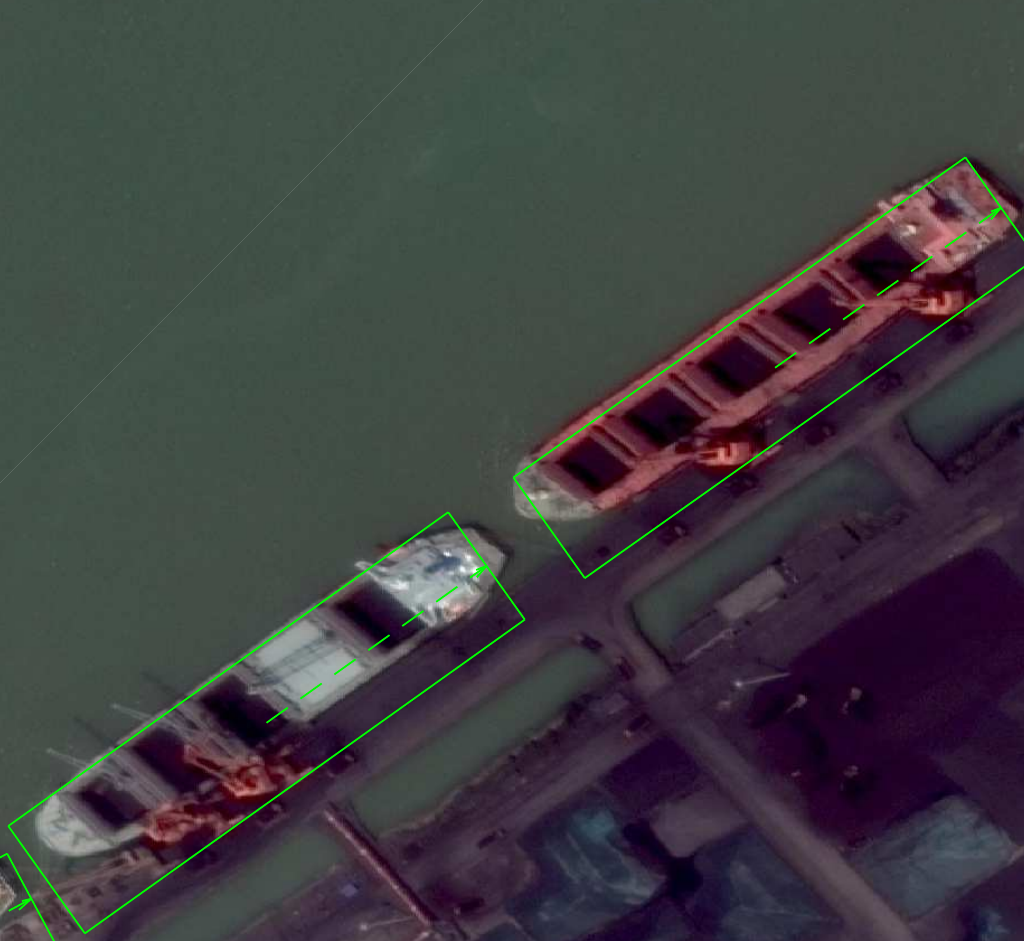}}
     \label{fig:result1}
   \subfloat[]{
     \includegraphics[width=0.18\linewidth,height=0.18\linewidth,trim={2.166cm 6.575cm 13.614cm 9.824cm},clip]{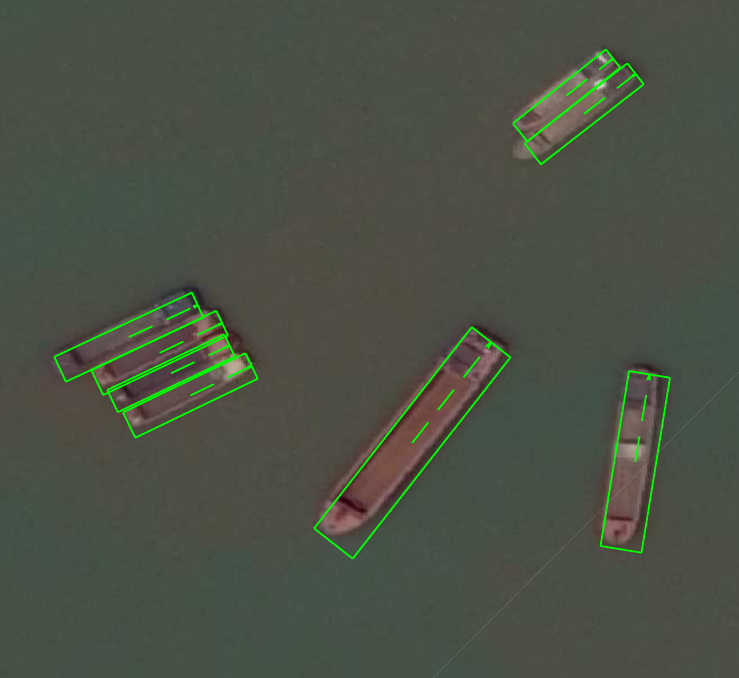}}
     \label{fig:result2}
   \subfloat[]{
     \includegraphics[width=0.18\linewidth,height=0.18\linewidth,trim={6.85cm 11.23cm 11.48cm 7.558cm},clip]{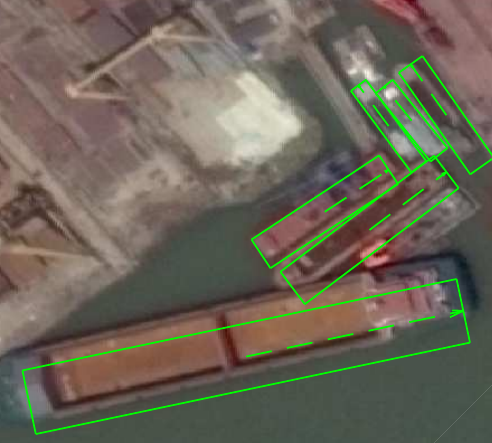}}
     \label{fig:result3}
\subfloat[]{
     \includegraphics[width=0.18\linewidth,height=0.18\linewidth,trim={0.077cm 10.945cm 11.951cm 1.702cm},clip]{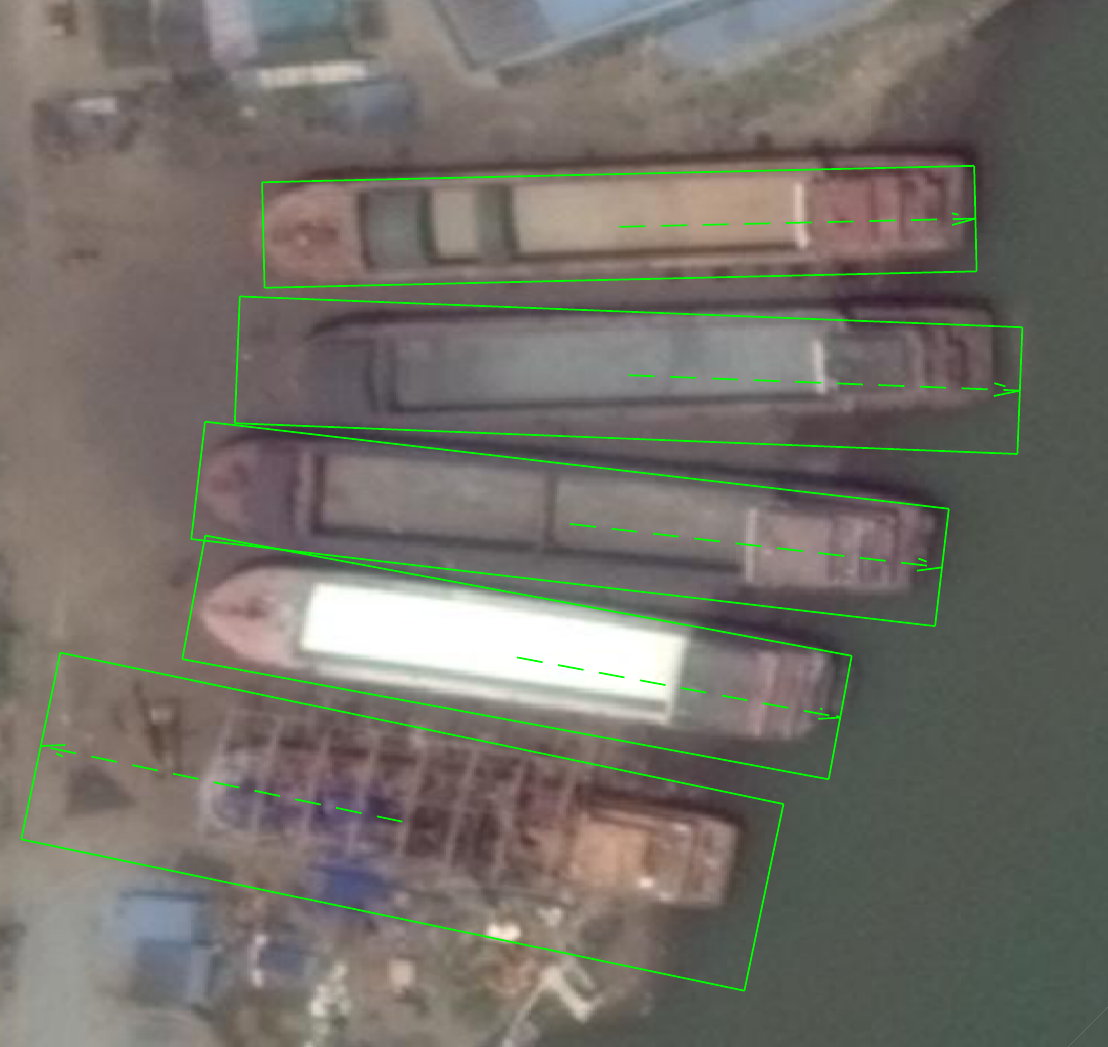}}
     \label{fig:result4}
\subfloat[]{
     \includegraphics[width=0.18\linewidth,height=0.18\linewidth,trim={0.696cm 5.26cm 15.042cm 10.946cm},clip]{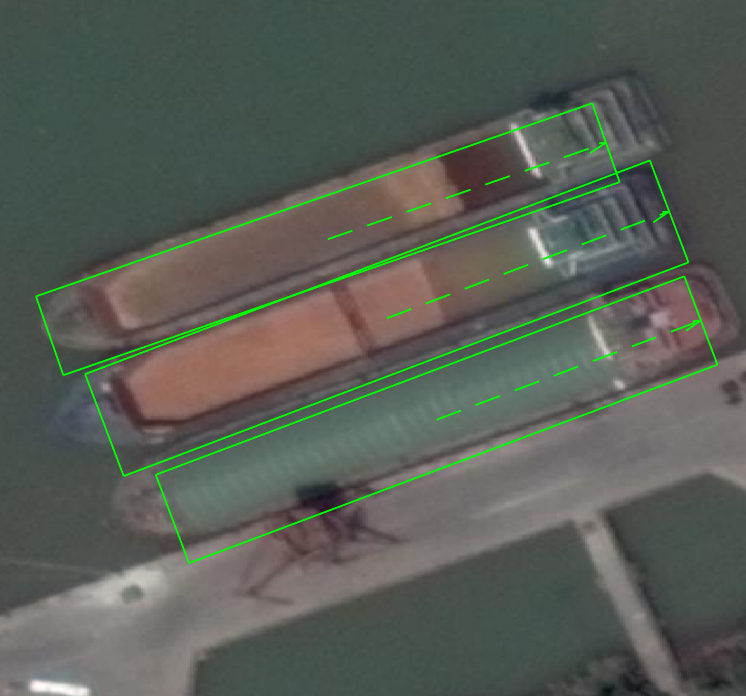}}
     \label{fig:result5}
\subfloat[]{
     \includegraphics[width=0.18\linewidth,height=0.18\linewidth]{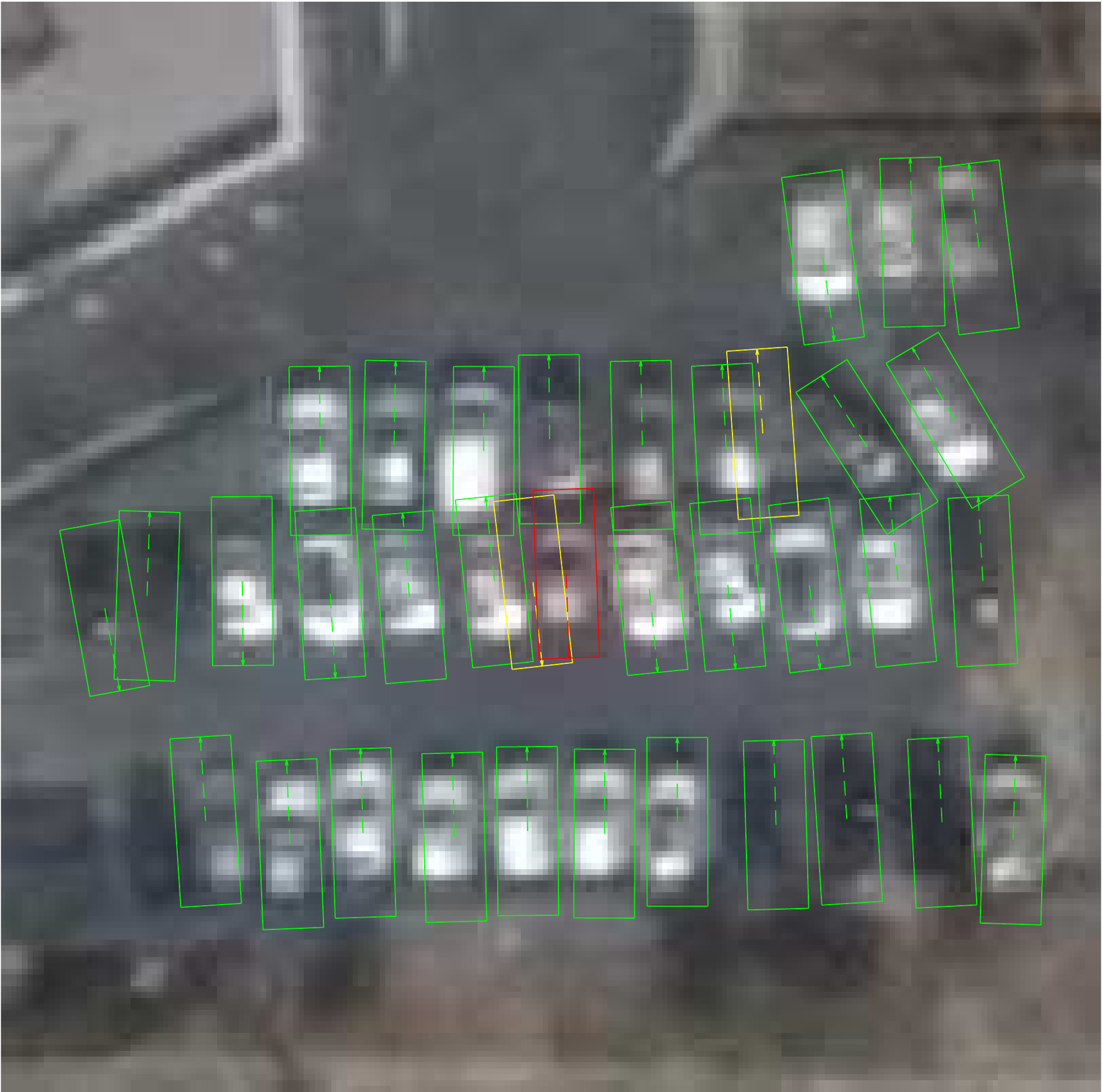}}
     \label{fig:result6}
   \subfloat[]{
     \includegraphics[width=0.18\linewidth,height=0.18\linewidth,trim={14.523cm 5.911cm 0.145cm 8.729cm},clip]{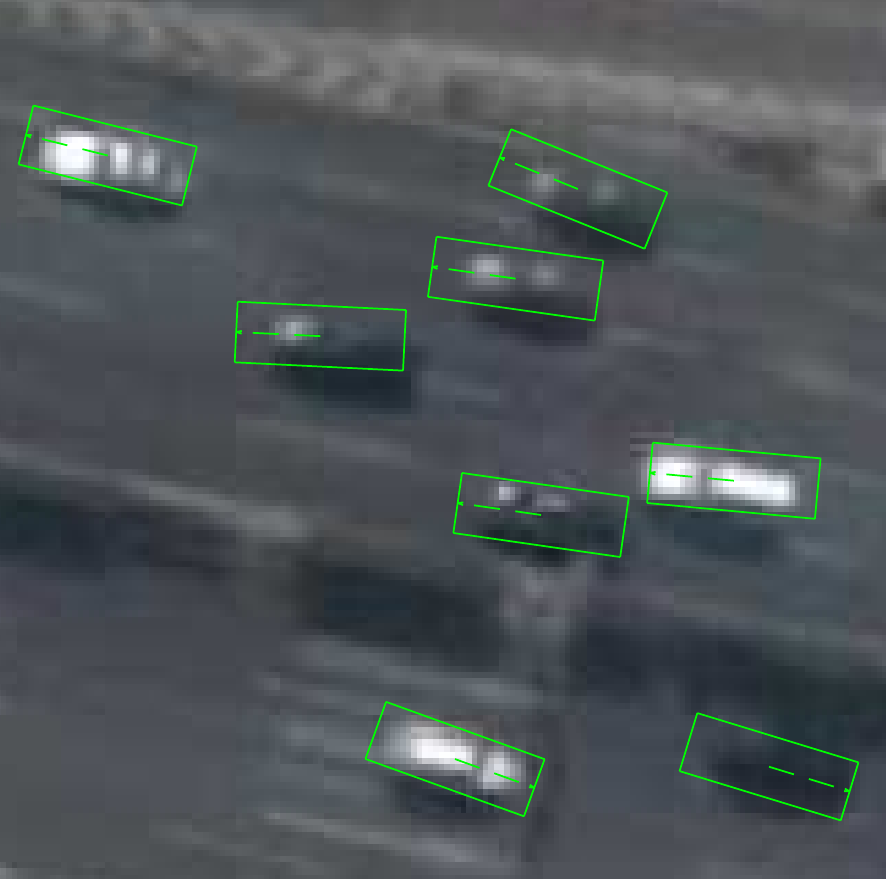}}
     \label{fig:result7}
   \subfloat[]{
     \includegraphics[width=0.18\linewidth,height=0.18\linewidth,trim={11.17cm 11.826cm 0.078cm 0cm},clip]{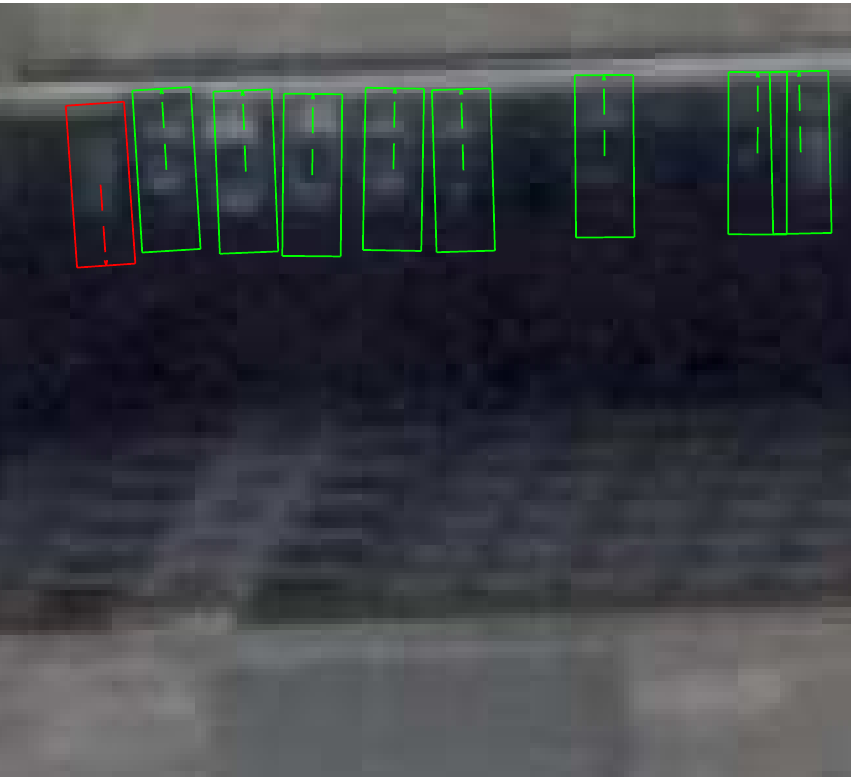}}
     \label{fig:result8}
\subfloat[]{
     \includegraphics[width=0.18\linewidth,height=0.18\linewidth]{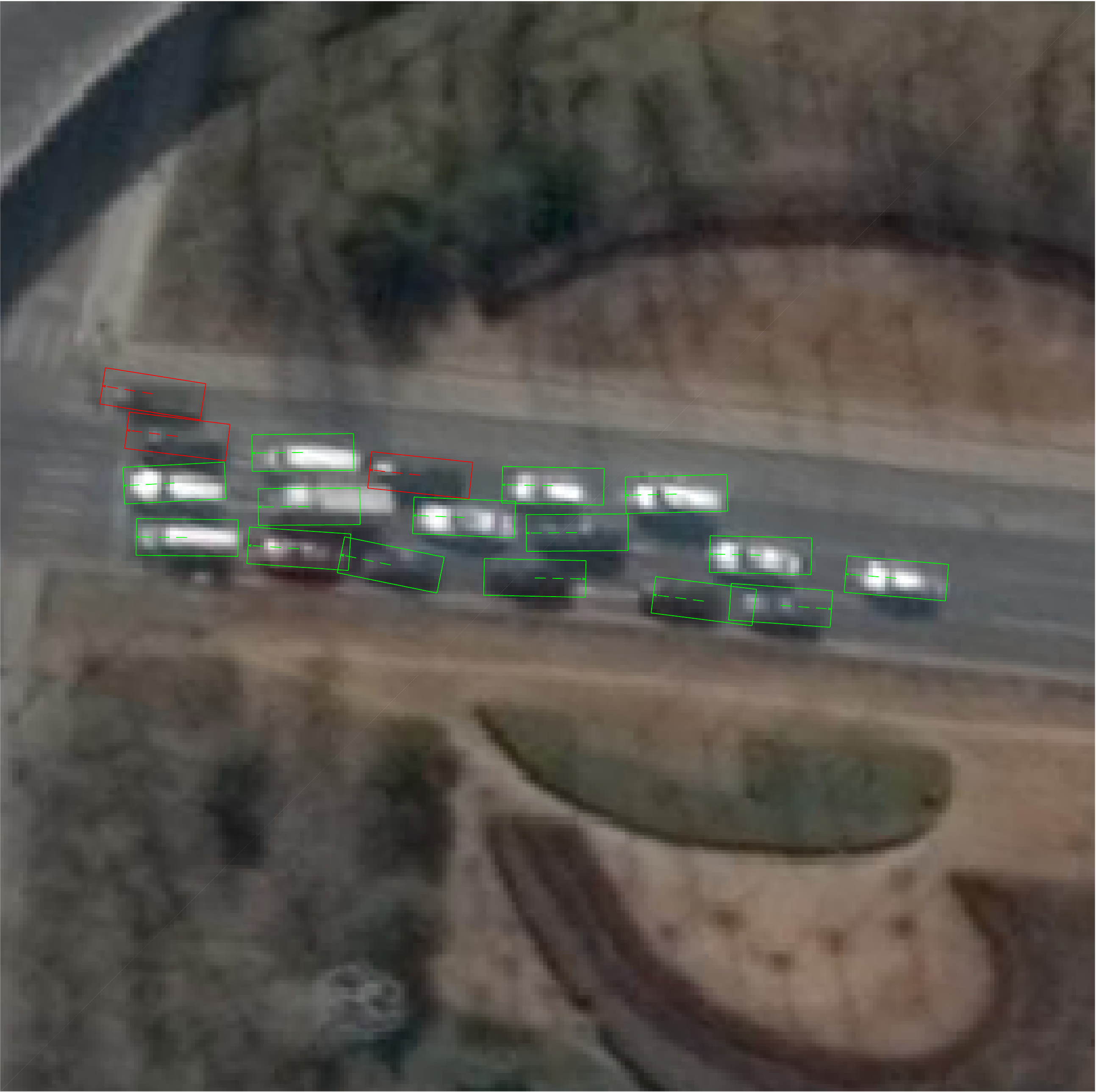}}
     \label{fig:result9}
\subfloat[]{
     \includegraphics[width=0.18\linewidth,height=0.18\linewidth]{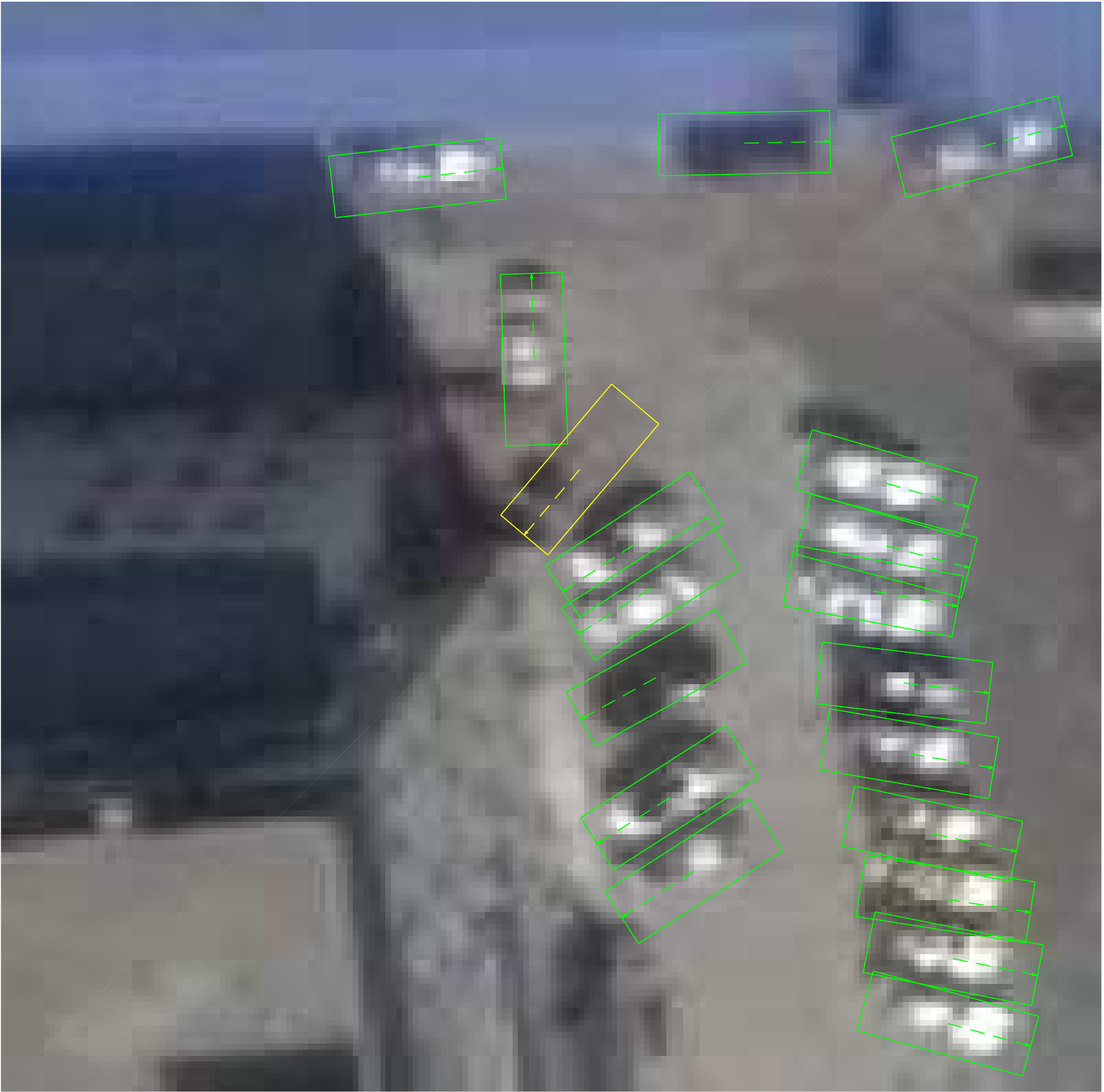}}
     \label{fig:result10}
\subfloat[]{
     \includegraphics[width=0.18\linewidth,height=0.18\linewidth]{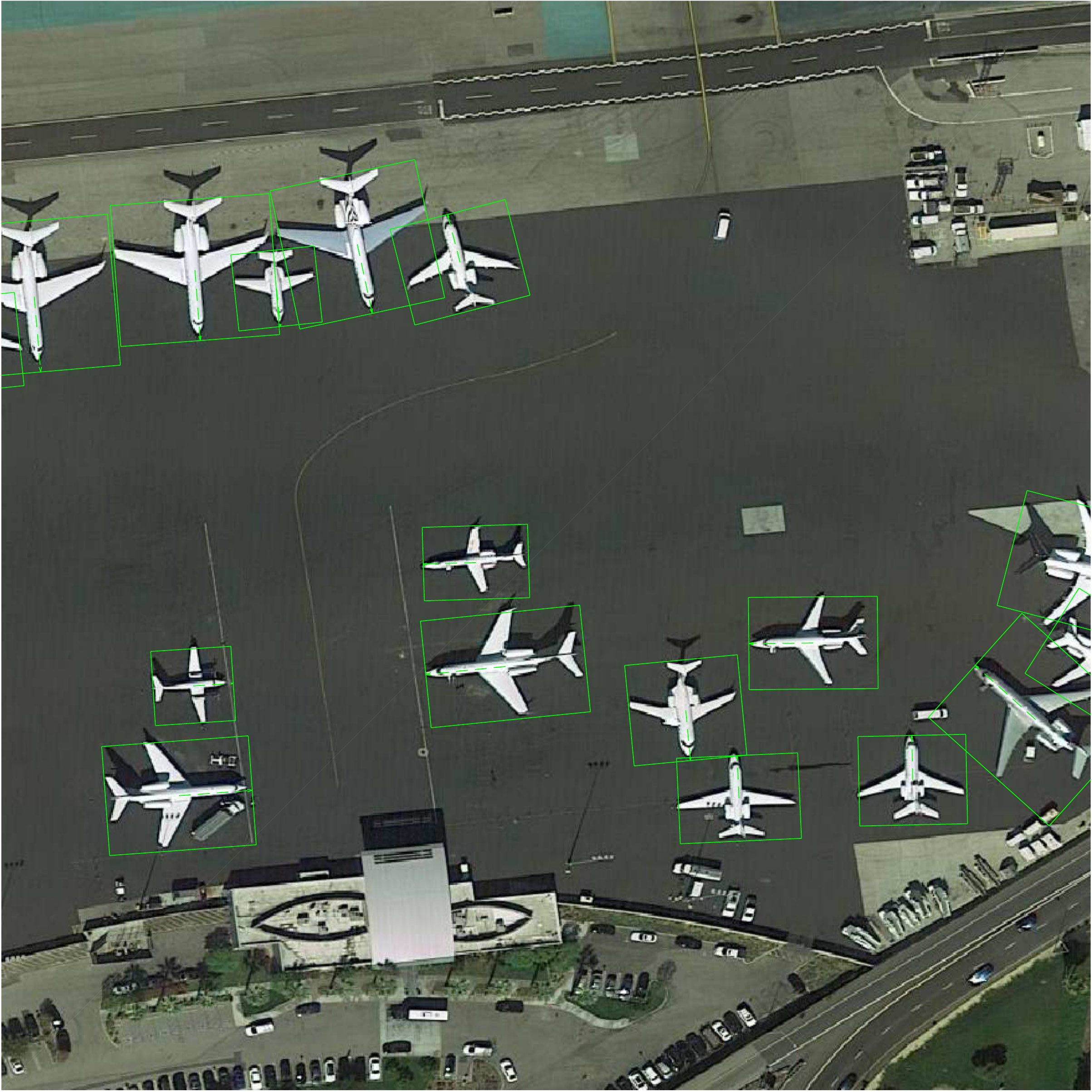}}
     \label{fig:result11}
   \subfloat[]{
     \includegraphics[width=0.18\linewidth,height=0.18\linewidth]{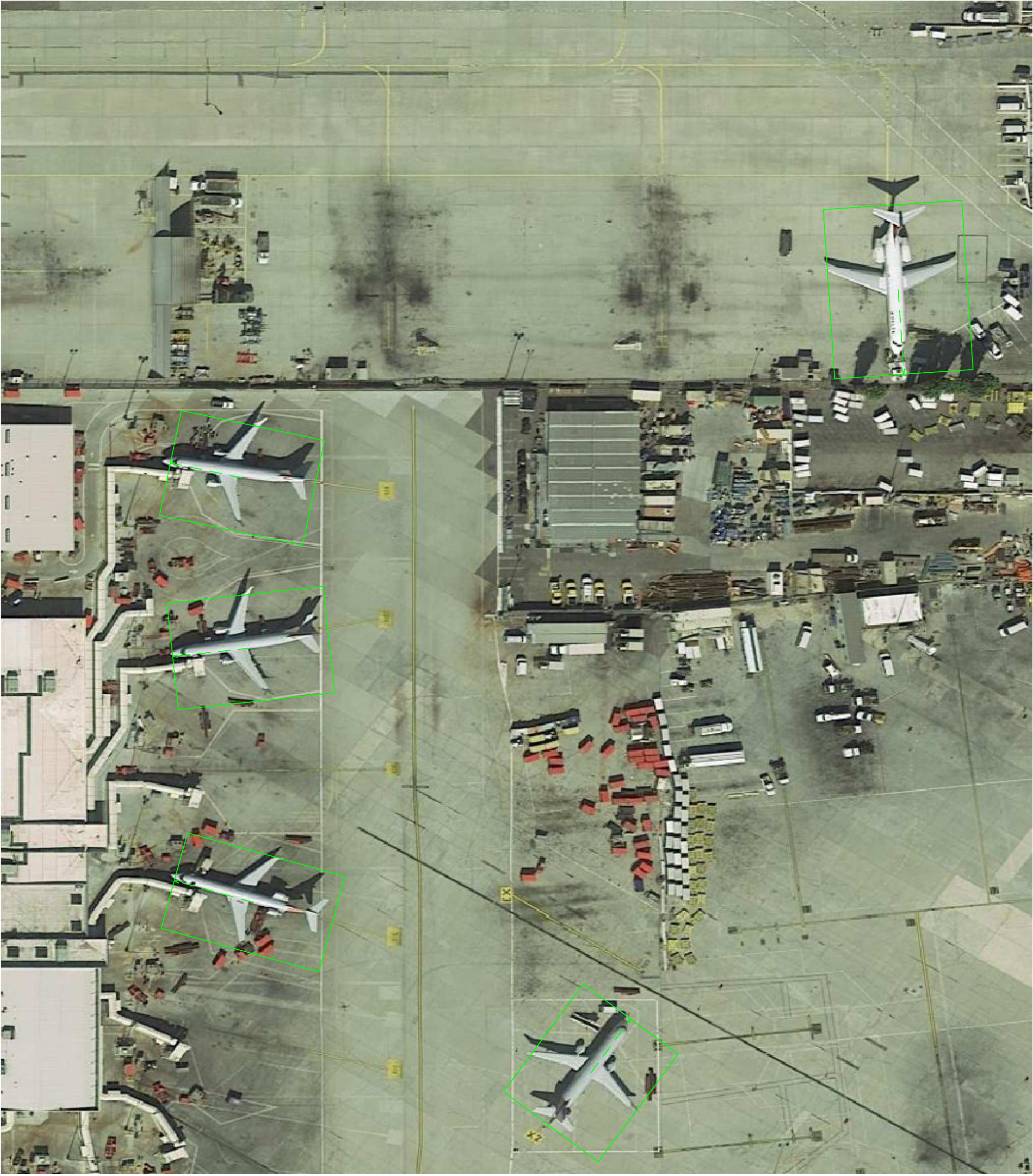}}
     \label{fig:result12}
   \subfloat[]{
     \includegraphics[width=0.18\linewidth,height=0.18\linewidth]{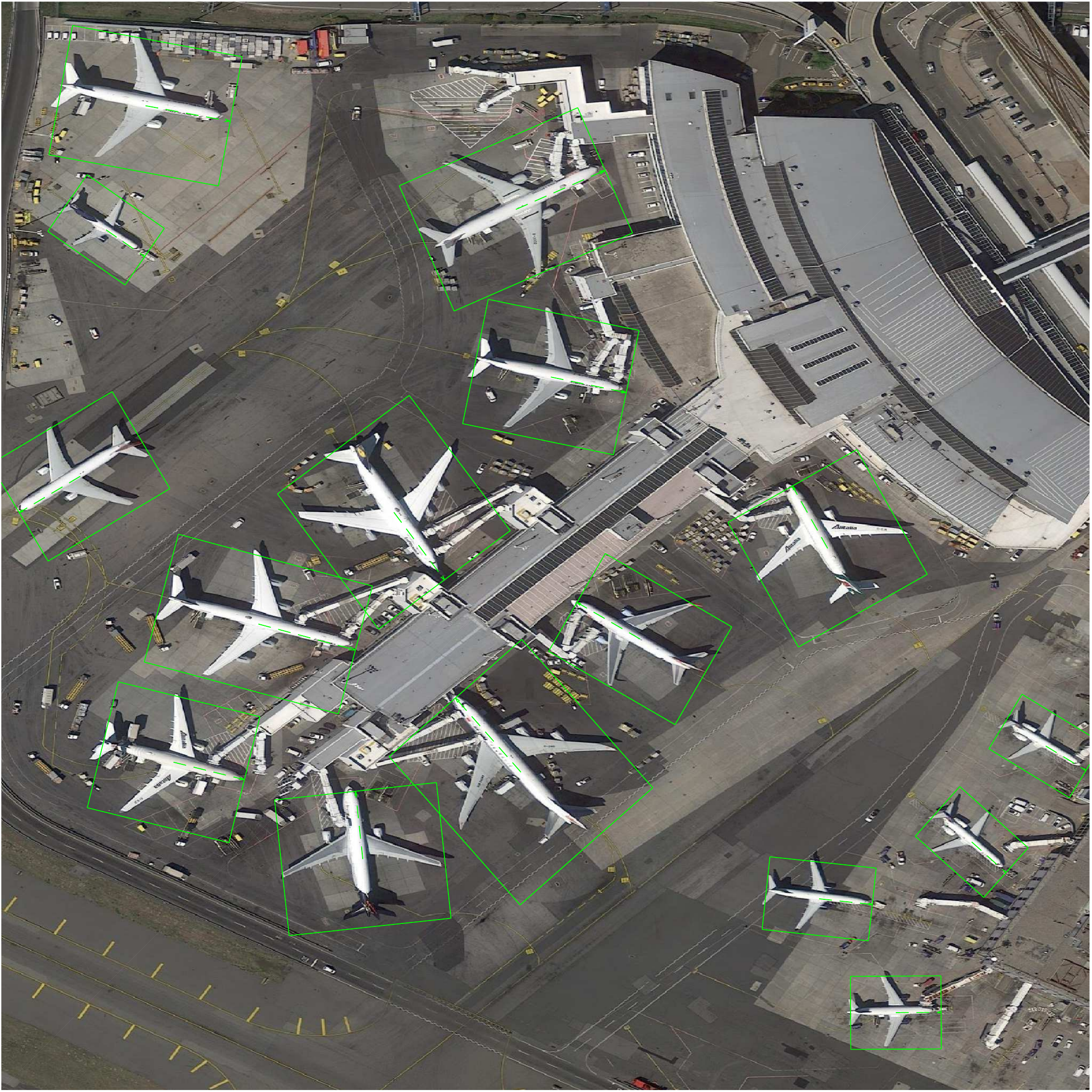}}
     \label{fig:result13}
\subfloat[]{
     \includegraphics[width=0.18\linewidth,height=0.18\linewidth]{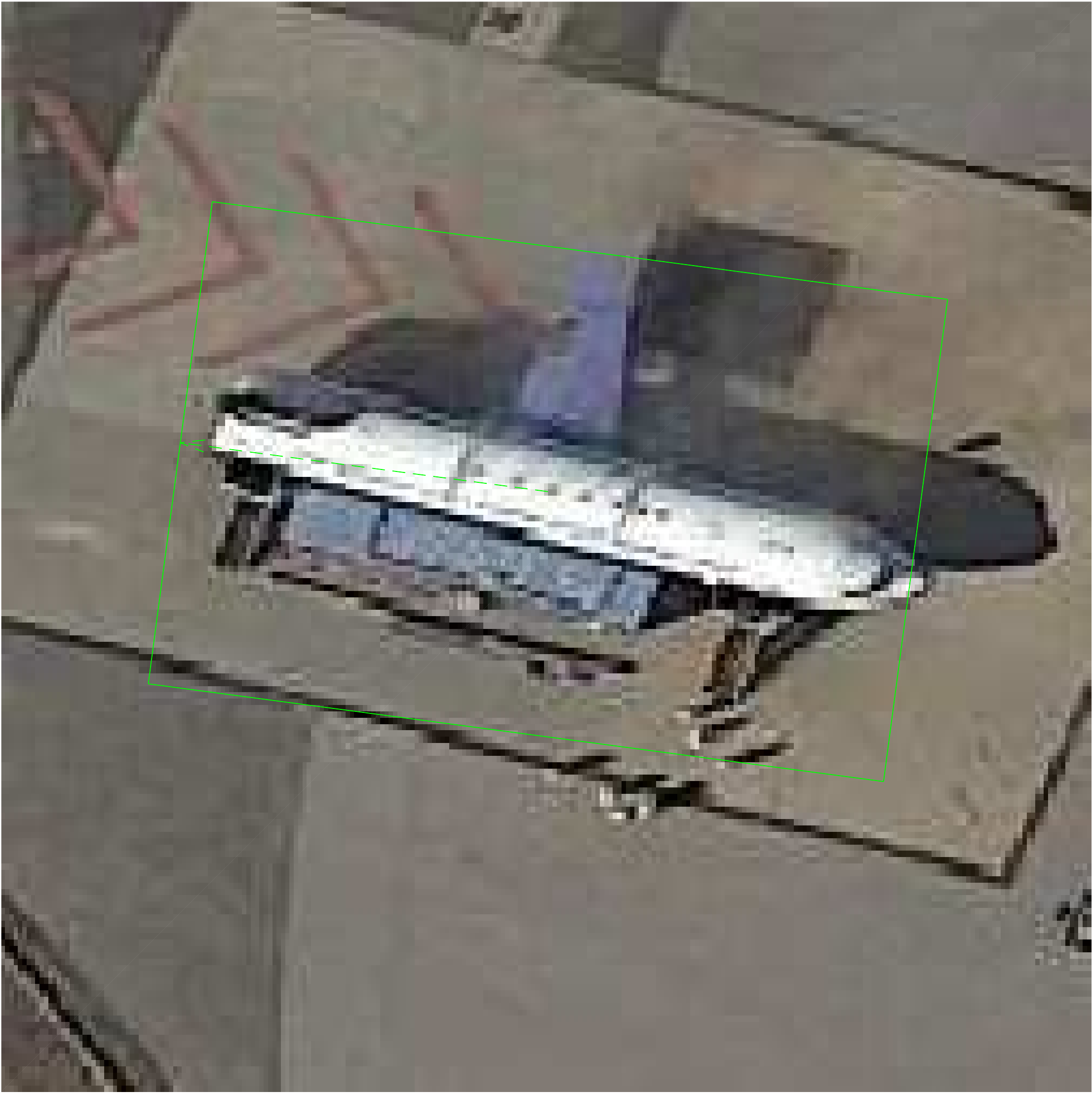}}
     \label{fig:result14}
\subfloat[]{
     \includegraphics[width=0.18\linewidth,height=0.18\linewidth]{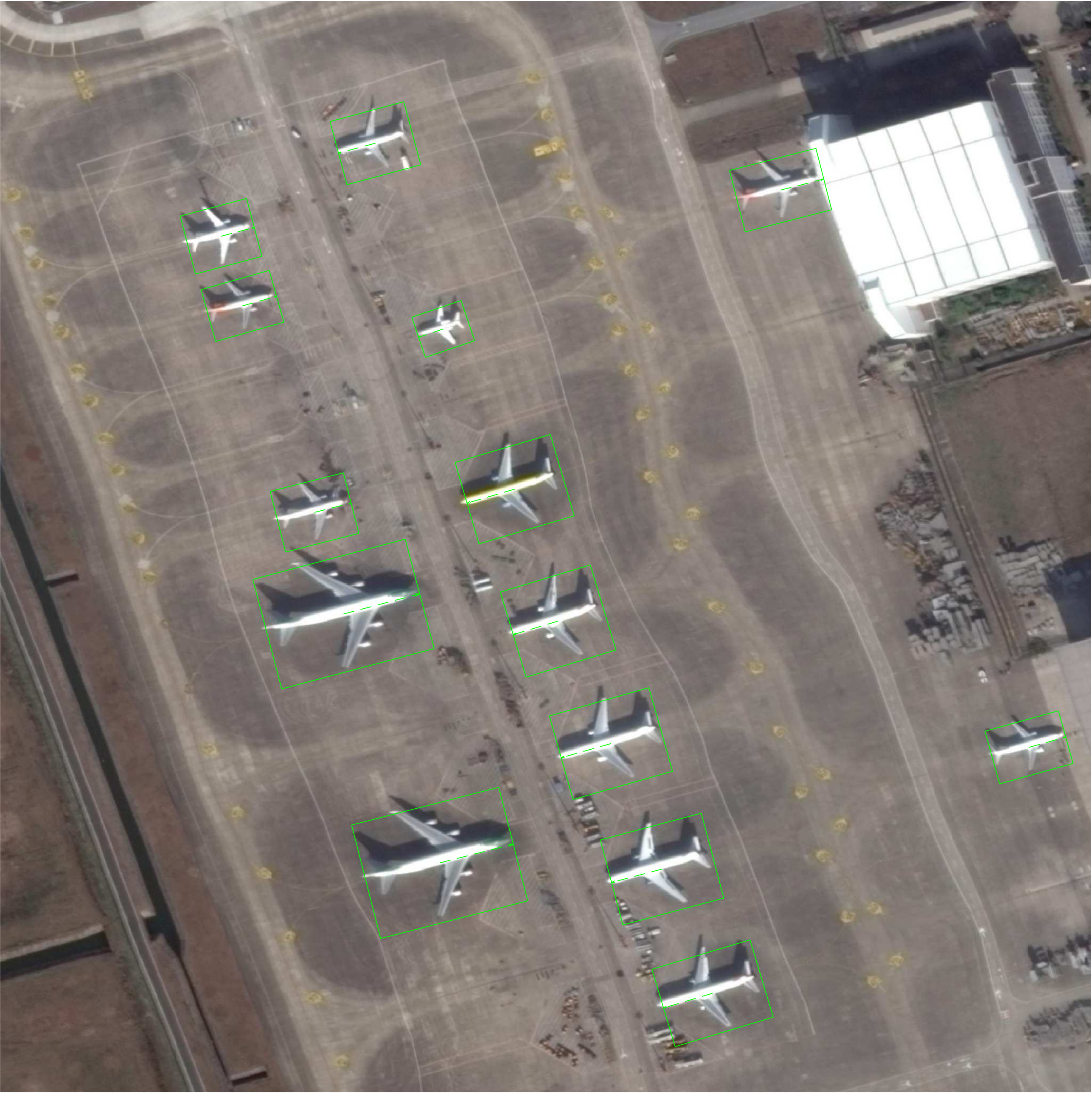}}
     \label{fig:result15}
\end{center}
   \caption{Detection results of DRBox. Ship detection results are shown in the first row, where DRBox works well in both open water region and port region. Vehicle detection results are shown in the second row, where DRBox successfully finds cars hidden in complex backgrounds and correctly indicates their head directions. Airplane detection results are shown in the third row, including an airplane that is under repairing.}
\label{fig:result}
\end{figure*}

More results of DRBox are shown in~\ref{fig:result}.  Ship detection in port region is more challengeable than in open water region, whereas DRBox works well on both situations. The vehicles are very difficult to detect due to small size and complex backgrounds. In our dataset, each car is around 20 pixels in length and 9 pixels in width. Fortunately, we find that DRBox successfully finds vehicles that hided in the shadows of tall buildings, or parked very close to each other. We also find that DRBox can not only output the locates of the cars, but also predict the head direction of each car, which is even a challenge task for human beings. The estimated direction of cars on road matches the prior knowledge that traffic always keeps to the right of the road. Airplanes with different sizes are succesfully detected, including an airplane that is under repairing. 

\begin{figure*}
\begin{center}
   \subfloat[]{
     \includegraphics[width=0.3\linewidth]{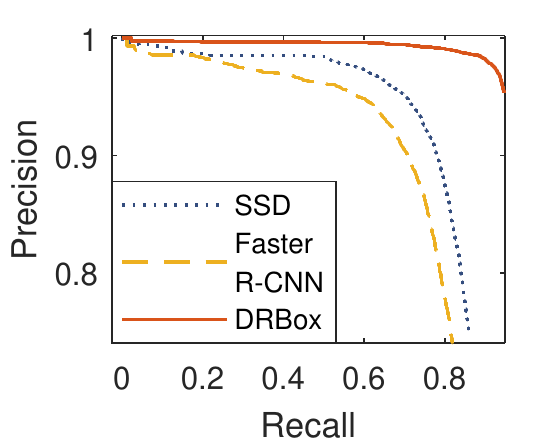}}
     \label{fig:roc1}
   \subfloat[]{
     \includegraphics[width=0.3\linewidth]{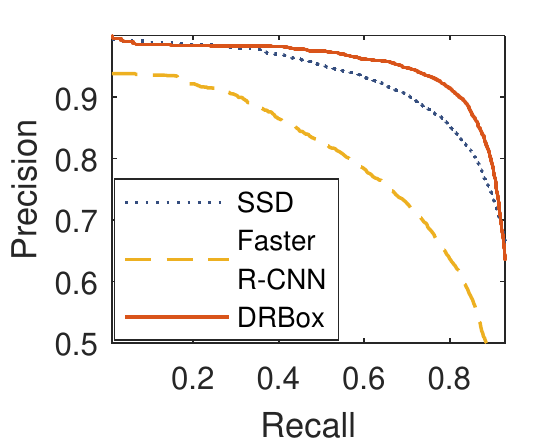}}
     \label{fig:roc2}
   \subfloat[]{
     \includegraphics[width=0.3\linewidth]{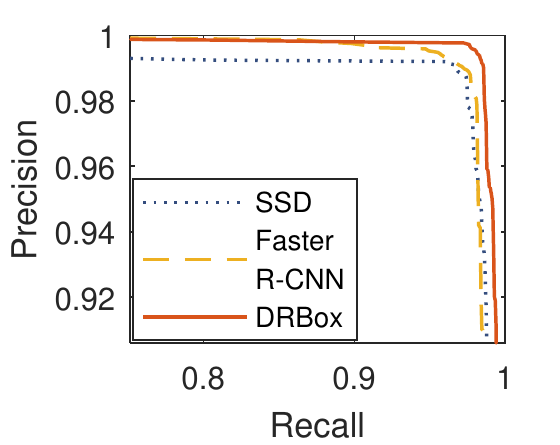}}
     \label{fig:roc3}
\end{center}
   \caption{Precision-recall curves of (a) ship detection results, (b) vehicle detection results and (c) airplane detection results. The performance of DRBox is the best in each detection task.}
\label{fig:roc}
\end{figure*}

Figure~\ref{fig:roc} shows precision-recall (P-R) curves of DRBox, SSD and Faster R-CNN. The recall ratio evaluates the ability of finding more targets in the image, while the precision evaluates the quality of predicting only the right object rather than containing many false alarms. The P-R curve of SSD and Faster R-CNN are always below the P-R curve of DRBox. DRBox has the best performance in this test. We further show BEP (Break-Even Point), AP (Average Precision) and mAP (mean Average Precision) of each method in Table~\ref{tab:map}. BEP is the point on P-R curve where precision equals recall, AP is the area below P-R curve, and mAP is the mean value of the APs on all object detection tasks. DRBox is always the best on all the indexes compared with SSD and Faster R-CNN.

\begin{table}
\begin{center}
\newcommand{\tabincell}[5]
{\begin{tabular}{@{}#1@{}}#2\end{tabular}}
\begin{tabular}{|l l|c c c|}
\hline
Method & Dataset & BEP(\%) & AP(\%) & mAP(\%)\\
\hline\hline
\multirow{3}{0.15\linewidth}{Faster R-CNN}&Ship&79.20&82.29&
\multirow{3}{*}{85.63} \\
&Vehicle&71.60&75.55& \\
&Airplane&98.07&99.06& \\
\hline
\multirow{3}{*}{SSD} &Ship&82.72&82.89& 
\multirow{3}{*}{89.68}\\
&Vehicle&83.13&87.59& \\
&Airplane&97.74&98.56& \\
\hline
\multirow{3}{*}{DRBox} &Ship&94.62&94.06&
\multirow{3}{*}{94.13}\\
&Vehicle&86.14&89.07& \\
&Airplane&98.62&99.28& \\
\hline
\end{tabular}
\end{center}
\caption{BEP, AP and mAP of Faster R-CNN, SSD and DRBox. DRBox outperforms the other two methods on all indexes listed in this table.}
\label{tab:map}
\end{table}

\subsection{Comparison of the robustness against rotation}
The robustness against rotation of a method involves two aspects: robustness against rotation of input images and robustness against rotation of objects. Robustness against rotation of input images means that a detection method should output the same results when the input image rotates arbitrarily. We define STD\_AP to quantify this ability. STD\_AP is the standard deviation of AP when the angle of the input image changes. STD\_AP is estimated by rotating the test images each 10 degrees, then calculate the AP values of the detection results, respectively, and the standard deviation of the APs. Robustness against rotation of objects means that the same object should always be successfully detected when its orientation changes. We define STD\_AS to quantify this ability. STD\_AS is the standard deviation of the average score for objects in different orientation angles, which is estimated by dividing the objects in testing dataset into different groups according to their angles, then calculate average score (softmax threshold) of each group, respectively, and the standard deviation over all groups. The two robustness evaluation methods are interrelated, except that STD\_AP has a bias towards the robustness on rotaton of backgrounds while STD\_AS has a bias towards the robustness on rotation of objects. The smaller STD\_AP and STD\_AS value indicate that the detection method is more robust against rotation.

STD\_AP and STD\_AS values of DRBox, SSD and Faster R-CNN are shown in Table~\ref{tab:std}. All the three methods are robust against rotation on airplane detection. However, Faster R-CNN is relatively not robust against rotation on ship and vehicle detection, SSD is not robust against rotation on ship detection. DRBox remains good scores in all tasks. 

\begin{figure*}
\begin{center}
	\includegraphics[width=0.9\linewidth]{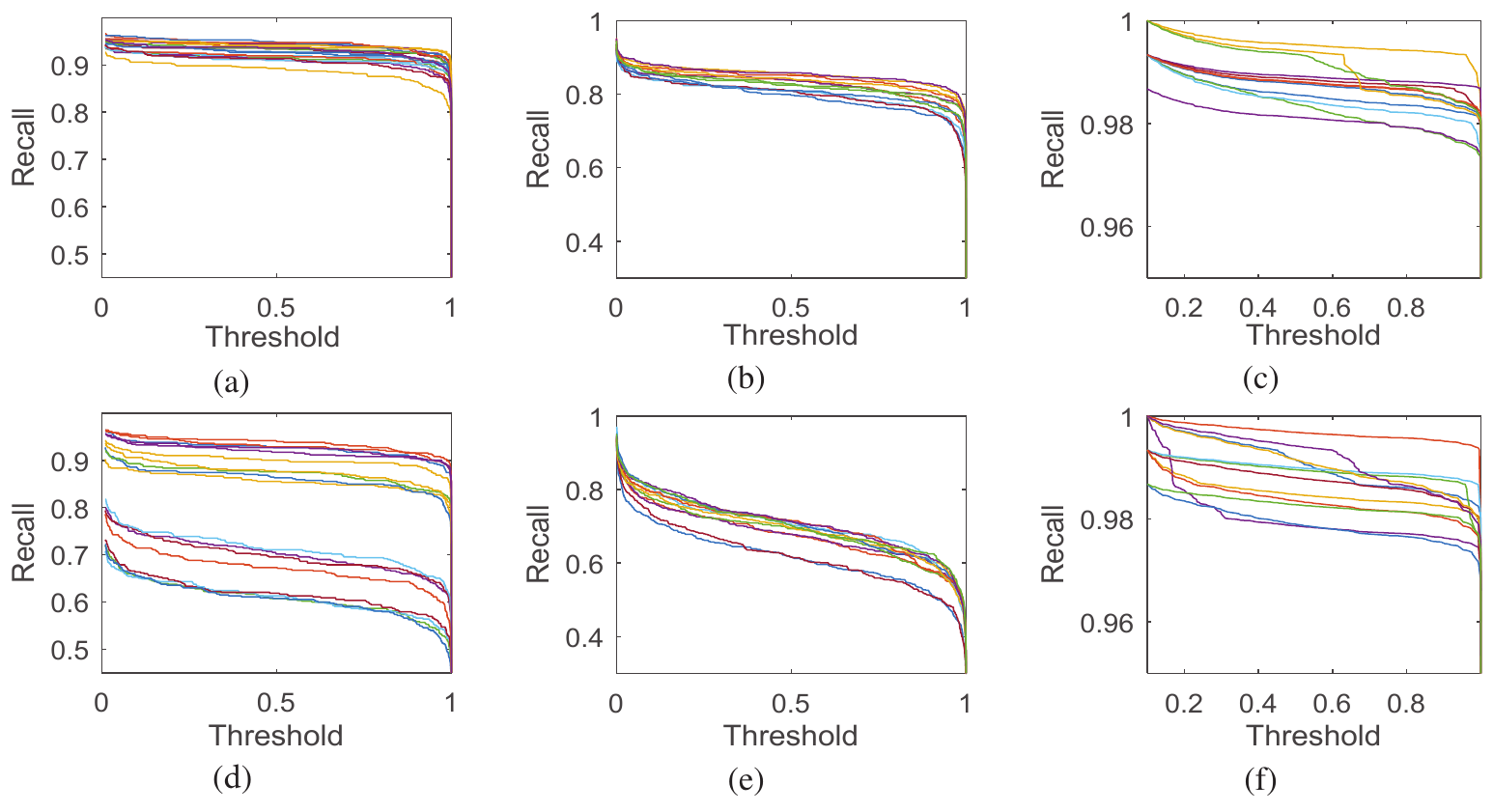}
\end{center}
   \caption{The curves of recall ratios on different softmax threshold values. In each sub-figure, different curves are for different object orientation angles. (a) (b) (c) are results of DRBox in ship detection task, vehicle detection task and airplane detection task, respectively; (d) (e) (f) are results of SSD in ship detection task, vehicle detection task and airplane detection task, respectively. The curves generated by DRBox are more concentrated, which indicates that DRBox is more robust to rotation of objects.}
\label{fig:multiangle2}
\end{figure*}

Further comparison between DRBox and SSD are demonstrated in Figure~\ref{fig:multiangle1} and Figure~\ref{fig:multiangle2}. Figure~\ref{fig:multiangle1} shows P-R recalls of the same input image but rotated to different angles, where the results of DRBox and SSD are plotted in shallow red and shallow blue colors, respectively. The curves generated by DRBox are more concentrated, which indicates that DRBox is more robust against rotation of input images compared with SSD. 

We calculate the recall ratio of targets in different angle scopes, respectively. Figure~\ref{fig:multiangle2} show the results. Each angle scope corresponds to one curve which shows the relationship between the softmax threshold and recall ratio. The performance of DRBox approximately remains the same for each angle scopes, whereas the performance of SSD shows strong instability when the angle of objects changes.

\section{Conclusions}
Robustness to rotation is very important on detection tasks of arbitrarily orientated objects. Existing detection algorithms uses bounding box to locate objects, which is a rotation variant structure. In this article, we replace the traditional bounding box with RBox and reconstruct deep CNN based detection frameworks with this new structure. The proposed detector, which is called DRBox, is rotation invariant due to its ability of estimating the orientation angles of objects. DRBox outperforms Faster R-CNN and SSD on object detection of satellite images.

DRBox is designed as a box-based method, whereas it is also possible to apply RBox into proposal based detection frameworks, e.g. R-FCN or Faster R-CNN. Training with RBox enforces the network to learn multi-scale local orientation information of the input image. We are looking forward of this interesting property to be used in other orientation sensitive tasks. 

\begin{table}
\begin{center}
\newcommand{\tabincell}[4]
{\begin{tabular}{@{}#1@{}}#2\end{tabular}}
\begin{tabular}{|l l|c c|}
\hline
Method & Dataset & STD\_AP(\%) & STD\_AS(\%) \\
\hline\hline
\multirow{3}{0.15\linewidth}{Faster R-CNN} &Ship&5.51&13.76\\
&Vehicle&7.21&7.82\\
&Airplane&0.28&0.17\\
\hline
\multirow{3}{*}{SSD} &Ship&5.97&13.14\\
&Vehicle&0.90&3.20\\
&Airplane&0.80&0.51\\
\hline
\multirow{3}{*}{DRBox} &Ship&0.88&1.56\\
&Vehicle&0.72&2.06 \\
&Airplane&0.51&0.41\\
\hline
\end{tabular}
\end{center}
\caption{Quantitative evaluation of robustness against rotation for Faster R-CNN, SSD and DRBox. STD\_AP evaluates robustness against rotation of input images. STD\_R evaluates robustness against rotation of objects. DRBox outperforms the other two methods except that Faster R-CNN is slightly more robust to rotation of airplane targets.}
\label{tab:std}
\end{table}

\captionsetup[subfigure]{labelformat=empty, labelsep=space, font=normal,farskip=-13pt}

\captionsetup[subfigure]{labelformat=parens, labelsep=space, font=normal}

\captionsetup[subfigure]{labelformat=parens, labelsep=space, font=normal}
\begin{figure*}
\begin{center}
   \subfloat[]{
     \includegraphics[width=0.3\linewidth, height=0.222\linewidth]{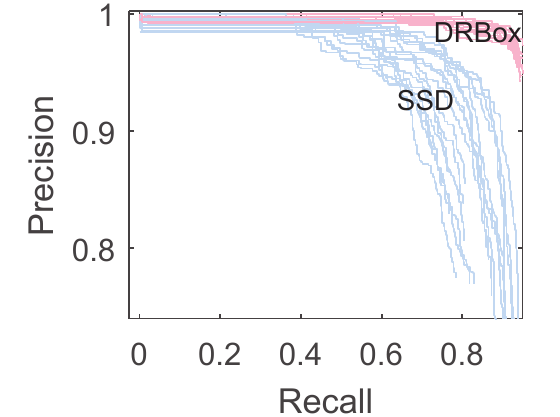}}
     \label{fig:multiangle1}
   \subfloat[]{
     \includegraphics[width=0.3\linewidth, height=0.222\linewidth]{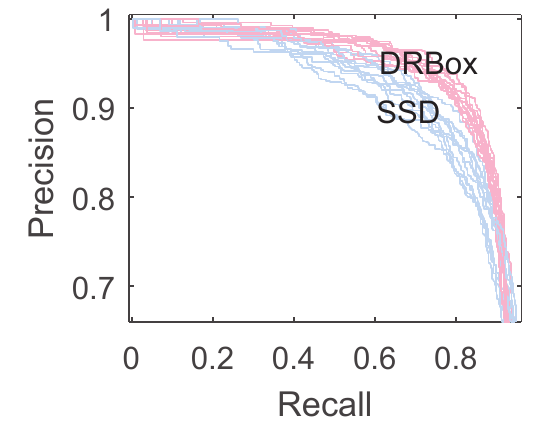}}
     \label{fig:multiangle4}
   \subfloat[]{
     \includegraphics[width=0.3\linewidth, height=0.222\linewidth]{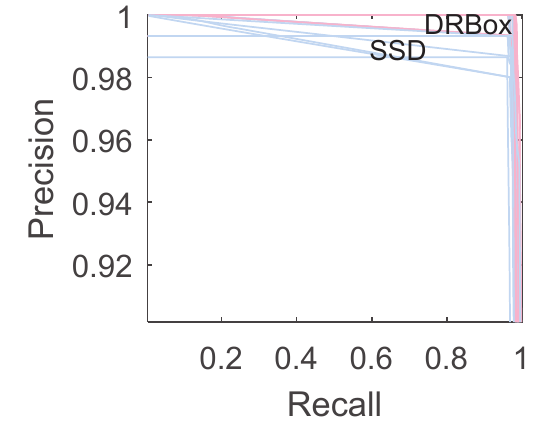}}
     \label{fig:multiangle7}
\end{center}
   \caption{Precision-recall curves when input images are rotated on multiple angles in (a) ship detection task, (b) vehicle detection task and (c) airplane detection task. The curves generated by SSD are in blue color, and the curves generated by DRBox are in red color. The curves generated by DRBox are more concentrated, which indicates that DRBox is more robust to rotation of input images.}
\label{fig:multiangle1}
\end{figure*}

\captionsetup[subfigure]{labelformat=parens, labelsep=space, font=normal}

\clearpage
\clearpage
{\small
\bibliographystyle{ieee}
\bibliography{egbib}

\begin{thebibliography}{10}\itemsep=-1pt

\bibitem{chen2017building}
X.~Chen, R.-X. Gong, L.-L. Xie, S.~Xiang, C.-L. Liu, and C.-H. Pan.
\newblock Building regional covariance descriptors for vehicle detection.
\newblock {\em IEEE Geoscience and Remote Sensing Letters}, 14(4):524--528,
  2017.

\bibitem{chen2014vehicle}
X.~Chen, S.~Xiang, C.-L. Liu, and C.-H. Pan.
\newblock Vehicle detection in satellite images by hybrid deep convolutional
  neural networks.
\newblock {\em IEEE Geoscience and remote sensing letters}, 11(10):1797--1801,
  2014.

\bibitem{chen2016vehicle2}
Z.~Chen, C.~Wang, H.~Luo, H.~Wang, Y.~Chen, C.~Wen, Y.~Yu, L.~Cao, and J.~Li.
\newblock Vehicle detection in high-resolution aerial images based on fast
  sparse representation classification and multiorder feature.
\newblock {\em IEEE Transactions on Intelligent Transportation Systems},
  17(8):2296--2309, 2016.

\bibitem{chen2016vehicle1}
Z.~Chen, C.~Wang, C.~Wen, X.~Teng, Y.~Chen, H.~Guan, H.~Luo, L.~Cao, and J.~Li.
\newblock Vehicle detection in high-resolution aerial images via sparse
  representation and superpixels.
\newblock {\em IEEE Transactions on Geoscience and Remote Sensing},
  54(1):103--116, 2016.

\bibitem{cheng2016learning}
G.~Cheng, P.~Zhou, and J.~Han.
\newblock Learning rotation-invariant convolutional neural networks for object
  detection in vhr optical remote sensing images.
\newblock {\em IEEE Transactions on Geoscience and Remote Sensing},
  54(12):7405--7415, 2016.

\bibitem{dai2016r}
J.~Dai, Y.~Li, K.~He, and J.~Sun.
\newblock R-fcn: Object detection via region-based fully convolutional
  networks.
\newblock In {\em Advances in neural information processing systems}, pages
  379--387, 2016.

\bibitem{diao2016efficient}
W.~Diao, X.~Sun, X.~Zheng, F.~Dou, H.~Wang, and K.~Fu.
\newblock Efficient saliency-based object detection in remote sensing images
  using deep belief networks.
\newblock {\em IEEE Geoscience and Remote Sensing Letters}, 13(2):137--141,
  2016.

\bibitem{felzenszwalb2010object}
P.~F. Felzenszwalb, R.~B. Girshick, D.~McAllester, and D.~Ramanan.
\newblock Object detection with discriminatively trained part-based models.
\newblock {\em IEEE transactions on pattern analysis and machine intelligence},
  32(9):1627--1645, 2010.

\bibitem{girshick2015fast}
R.~Girshick.
\newblock Fast r-cnn.
\newblock In {\em Proceedings of the IEEE international conference on computer
  vision}, pages 1440--1448, 2015.

\bibitem{girshick2014rich}
R.~Girshick, J.~Donahue, T.~Darrell, and J.~Malik.
\newblock Rich feature hierarchies for accurate object detection and semantic
  segmentation.
\newblock In {\em Proceedings of the IEEE conference on computer vision and
  pattern recognition}, pages 580--587, 2014.

\bibitem{han2015object}
J.~Han, D.~Zhang, G.~Cheng, L.~Guo, and J.~Ren.
\newblock Object detection in optical remote sensing images based on weakly
  supervised learning and high-level feature learning.
\newblock {\em IEEE Transactions on Geoscience and Remote Sensing},
  53(6):3325--3337, 2015.

\bibitem{he2017mask}
K.~He, G.~Gkioxari, P.~Doll{\'a}r, and R.~Girshick.
\newblock Mask r-cnn.
\newblock {\em arXiv preprint arXiv:1703.06870}, 2017.

\bibitem{he2014spatial}
K.~He, X.~Zhang, S.~Ren, and J.~Sun.
\newblock Spatial pyramid pooling in deep convolutional networks for visual
  recognition.
\newblock In {\em European Conference on Computer Vision}, pages 346--361.
  Springer, 2014.

\bibitem{jiang2015deep}
Q.~Jiang, L.~Cao, M.~Cheng, C.~Wang, and J.~Li.
\newblock Deep neural networks-based vehicle detection in satellite images.
\newblock In {\em Bioelectronics and Bioinformatics (ISBB), 2015 International
  Symposium on}, pages 184--187. IEEE, 2015.

\bibitem{lin2016feature}
T.-Y. Lin, P.~Doll{\'a}r, R.~Girshick, K.~He, B.~Hariharan, and S.~Belongie.
\newblock Feature pyramid networks for object detection.
\newblock {\em arXiv preprint arXiv:1612.03144}, 2016.

\bibitem{liu2016ssd}
W.~Liu, D.~Anguelov, D.~Erhan, C.~Szegedy, S.~Reed, C.-Y. Fu, and A.~C. Berg.
\newblock Ssd: Single shot multibox detector.
\newblock In {\em European conference on computer vision}, pages 21--37.
  Springer, 2016.

\bibitem{long2017accurate}
Y.~Long, Y.~Gong, Z.~Xiao, and Q.~Liu.
\newblock Accurate object localization in remote sensing images based on
  convolutional neural networks.
\newblock {\em IEEE Transactions on Geoscience and Remote Sensing},
  55(5):2486--2498, 2017.

\bibitem{redmon2016you}
J.~Redmon, S.~Divvala, R.~Girshick, and A.~Farhadi.
\newblock You only look once: Unified, real-time object detection.
\newblock In {\em Proceedings of the IEEE Conference on Computer Vision and
  Pattern Recognition}, pages 779--788, 2016.

\bibitem{ren2015faster}
S.~Ren, K.~He, R.~Girshick, and J.~Sun.
\newblock Faster r-cnn: Towards real-time object detection with region proposal
  networks.
\newblock In {\em Advances in neural information processing systems}, pages
  91--99, 2015.

\bibitem{vsevo2016convolutional}
I.~{\v{S}}evo and A.~Avramovi{\'c}.
\newblock Convolutional neural network based automatic object detection on
  aerial images.
\newblock {\em IEEE Geoscience and Remote Sensing Letters}, 13(5):740--744,
  2016.

\bibitem{sommer2017fast}
L.~W. Sommer, T.~Schuchert, and J.~Beyerer.
\newblock Fast deep vehicle detection in aerial images.
\newblock In {\em Applications of Computer Vision (WACV), 2017 IEEE Winter
  Conference on}, pages 311--319. IEEE, 2017.

\bibitem{sun2012automatic}
H.~Sun, X.~Sun, H.~Wang, Y.~Li, and X.~Li.
\newblock Automatic target detection in high-resolution remote sensing images
  using spatial sparse coding bag-of-words model.
\newblock {\em IEEE Geoscience and Remote Sensing Letters}, 9(1):109--113,
  2012.

\bibitem{tuermer2013airborne}
S.~Tuermer, F.~Kurz, P.~Reinartz, and U.~Stilla.
\newblock Airborne vehicle detection in dense urban areas using hog features
  and disparity maps.
\newblock {\em IEEE Journal of Selected Topics in Applied Earth Observations
  and Remote Sensing}, 6(6):2327--2337, 2013.

\bibitem{uijlings2013selective}
J.~R. Uijlings, K.~E. Van De~Sande, T.~Gevers, and A.~W. Smeulders.
\newblock Selective search for object recognition.
\newblock {\em International journal of computer vision}, 104(2):154--171,
  2013.

\bibitem{wan2017affine}
L.~Wan, L.~Zheng, H.~Huo, and T.~Fang.
\newblock Affine invariant description and large-margin dimensionality
  reduction for target detection in optical remote sensing images.
\newblock {\em IEEE Geoscience and Remote Sensing Letters}, 2017.

\bibitem{zhang2016weakly}
F.~Zhang, B.~Du, L.~Zhang, and M.~Xu.
\newblock Weakly supervised learning based on coupled convolutional neural
  networks for aircraft detection.
\newblock {\em IEEE Transactions on Geoscience and Remote Sensing},
  54(9):5553--5563, 2016.

\bibitem{zhou2017orn}
Y.~Zhou, Q.~Ye, Q.~Qiu, and J.~Jiao.
\newblock Oriented response networks.
\newblock In {\em Proceedings of the IEEE Conference on Computer Vision and
  Pattern Recognition}, July 2017.

\bibitem{zitnick2014edge}
C.~L. Zitnick and P.~Doll{\'a}r.
\newblock Edge boxes: Locating object proposals from edges.
\newblock In {\em European Conference on Computer Vision}, pages 391--405.
  Springer, 2014.

\end{thebibliography}
}

\end{document}